\documentclass[letterpaper]{article}
\usepackage{aviotpaper}
\noreserved
\usepackage[hyphens]{url}
\usepackage{graphicx}
\usepackage{natbib}
\usepackage{caption}
\usepackage{amsmath}
\usepackage{amssymb}
\usepackage{algorithm}
\usepackage{algorithmic}

\usepackage{newfloat}
\usepackage{listings}
\DeclareCaptionStyle{ruled}{labelfont=normalfont,labelsep=colon,strut=off}
\floatstyle{ruled}
\newfloat{listing}{tb}{lst}{}
\floatname{listing}{Listing}

\usepackage{booktabs}
\usepackage{multirow}
\usepackage[hidelinks]{hyperref}
\pdftrailerid{}
\hypersetup{
  pdfhighlight=/N,
  urlbordercolor={0 0 0},
  pdftitle={Aggregating Visual Information with Optimal Transport for VideoLM Token Compression},
  pdfauthor={Wenti Yin, Xiaotian Han, Junyuan Shang, Yuchen Ding, Shuohuan Wang, Dianhai Yu, Changxin Gao, Nong Sang},
  pdfsubject={},
  pdfcreator={},
  pdfproducer={}
}
\title{Aggregating Visual Information with Optimal Transport for VideoLM Token Compression}
\author{
  Wenti Yin\textsuperscript{\rm 1,2},
  Xiaotian Han\textsuperscript{\rm 2}\thanks{Project Lead.},
  Junyuan Shang\textsuperscript{\rm 2},
  Yuchen Ding\textsuperscript{\rm 2},\\
  Shuohuan Wang\textsuperscript{\rm 2},
  Dianhai Yu\textsuperscript{\rm 2},
  Changxin Gao\textsuperscript{\rm 1},
  Nong Sang\textsuperscript{\rm 1}\thanks{Corresponding author.}
}
\affiliations{
  \textsuperscript{\rm 1}Key Laboratory of Image Processing and Intelligent Control,
  School of Artificial Intelligence and Automation,\\
  Huazhong University of Science and Technology\\
  \textsuperscript{\rm 2}Baidu, Inc\\
  \{yinwt, cgao, nsang\}@hust.edu.cn\\
  \{hanxiaotian, shangjunyuan, dingyuchen, wangshuohuan, yudianhai\}@baidu.com
}

\begin{document}

\maketitle

\begin{abstract}
Video language models process videos as dense visual-token sequences with substantial representational redundancy. 
Compressing these sequences is therefore essential for reducing the visual-token burden
on language-model decoding. The central challenge is to preserve visual information dispersed across frames under 
such compression. To this end, we introduce Aggregating Visual Information
with Optimal Transport (AVIOT), which casts video token
compression as transporting a dense empirical measure of frame observations
onto a compact target measure. The resulting source-to-target coupling
induces a distribution over source observations for each target support, directly specifying how 
the compressed video representation is constructed. 
We further adapt this construction along task and spatial axes. Question conditioning
modulates the transport cost between source frames and target supports, while
influencing how many supports are allocated to each temporal segment, thereby
directing representation capacity toward question-relevant content.
At multiple spatial granularities, AVIOT computes region-specific temporal transport plans and 
adaptively fuses the representations they yield, allowing different regions within the same compact representation to draw from different moments. 
Evaluations across varying compression ratios show that AVIOT
matches or outperforms the uncompressed baseline on multiple video-understanding
benchmarks while retaining strong performance at higher compression ratios. 
Code is available at
\href{https://github.com/ernie-research/AVIOT}{\nolinkurl{https://github.com/ernie-research/AVIOT}}.
\end{abstract}

\section{Introduction}
\label{sec:introduction}

Video language models (VideoLMs) have extended multimodal reasoning from static images to
dynamic visual content, enabling models to understand actions, events, and
temporal relations within videos \cite{zhang2025llavavideo}. Current VideoLMs
encode sampled frames as spatial grids of patch tokens and place the resulting
sequence in the language-model context. As temporal coverage expands, however,
every additional frame contributes an entire token grid, while slowly changing
scenes, objects, and backgrounds create substantial repetition. The visual
sequence therefore grows linearly with the number of sampled frames and can
dominate the computation and memory required for multimodal inference
\cite{shen2025longvu,fu2025framefusion}. Video token compression is thus
essential for reducing this burden while retaining information needed for video
understanding.

\begin{figure}[t]
    \centering
    \includegraphics[width=\columnwidth]{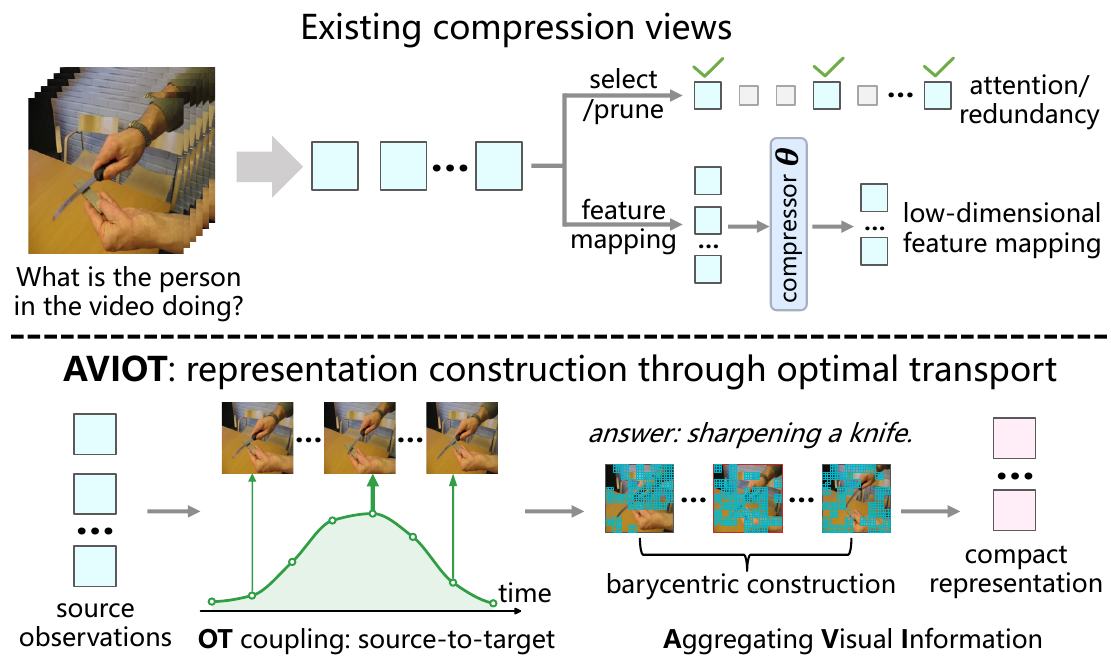}
    \caption{Motivation for AVIOT. Common compression
    paradigms reduce source observations through selection or pruning, or synthesize
    compact outputs through learned feature mapping. AVIOT constructs the output set
    through a joint source-to-target optimal-transport coupling, organizing how information
    distributed across frames is represented collectively by the compact supports.}
    \label{fig:motivation}
\end{figure}

Existing compression methods broadly follow two routes. One operates directly
on the source observations, selecting salient frames or tokens according to
attention, redundancy, or query relevance
\cite{chen2024fastv,shen2025longvu,zhang2025qframe}, or consolidating related
content through clustering and anchor-based merging
\cite{jin2024chatunivi,shang2025llavaprumerge,fu2025framefusion}. The other
learns a compact visual bottleneck, using adaptive tokenization or latent
queries to distill dense features into a small set of outputs
\cite{ryoo2021tokenlearner,alayrac2022flamingo}. Both routes have proved
effective, yet source-side reduction organizes compact outputs around retained
observations or source-derived anchors, while standard learned readouts
synthesize each output from an output-specific weighting of source features.
Such weighting specifies how each output is formed, but does not constrain how
the dense observations are represented across the output set as a whole. A
more comprehensive formulation should treat representation construction itself
as the compression problem, aggregating information distributed across frames
into compact supports that jointly account for the dense source distribution,
as illustrated in Fig.~\ref{fig:motivation}.

Optimal transport (OT) provides a natural mathematical framework for this
formulation: a coupling jointly allocates mass between source and target
measures \cite{cuturi2013sinkhorn}. Its prescribed source and target marginals
couple all target-wise assignments within a single plan, explicitly relating
the compact support set to the source measure as a whole. We introduce
\emph{Aggregating Visual Information with Optimal Transport} (AVIOT), an
optimal-transport framework for video token compression that transports a dense
empirical measure of frame observations onto a compact target measure with a
prescribed number of supports. An entropy-regularized transport coupling defines
their joint source-to-target correspondence. For each target support, its
normalized incoming mass defines a distribution over source observations; barycentric
projection uses this distribution to refine the support descriptor, and the
same distribution constructs its full spatial features. Each target support
can integrate visual information dispersed across multiple moments;
collectively, the supports account for the dense source distribution while
retaining a complete spatial layout.

AVIOT further adapts this representation construction along task and spatial
axes. Along the task axis, question conditioning modulates the transport cost
to change how source frames are associated with target supports. The question
also influences how many supports are allocated to each contiguous temporal
segment by combining segment-level relevance with the deviation of a
preliminary transport response from uniformity, distributing representation
capacity according to both signals. Along the spatial axis, AVIOT estimates
temporal correspondence under complementary global, medium, and local spatial
contexts. The resulting plans yield aligned representations with shared target
indices and spatial layout, which are adaptively fused at each target support
and spatial position. Different regions within the same compact representation
can thereby draw from different source moments while retaining global context
and a complete spatial layout.

Our main contributions are:
\begingroup
\renewcommand{\labelitemi}{$\bullet$}
\begin{itemize}
    \item We formulate video token compression as OT representation
    construction, transporting a dense empirical measure of frame observations
    onto a compact target measure whose target-conditional source distributions
    directly construct the compressed video representation.
    \item We adapt the same source-to-target construction along task and spatial
    axes through question-conditioned transport and transport across spatial
    granularities, enabling question-relevant capacity allocation and
    region-specific temporal correspondence while preserving spatial layout.
    \item We evaluate AVIOT across video-understanding benchmarks and
    compression ratios, matching or surpassing the uncompressed backbone on
    multiple benchmarks while remaining competitive at higher compression
    ratios.
\end{itemize}
\endgroup

\section{Related Work}
\label{sec:related_work}

\subsection{Video Language Models}

Video language models extend image-based multimodal large language models
(MLLMs) to temporal understanding by
projecting encoded video features into the language-model context and training
with multimodal instructions. Video-ChatGPT combines a video-adapted visual
encoder with video instruction data \cite{maaz2024videochatgpt}; Video-LLaVA
aligns image and video representations \cite{lin2024videollava}; and Chat-UniVi
uses dynamic visual tokens across both modalities \cite{jin2024chatunivi}.
Generalist systems such as LLaVA-OneVision (LLaVA-OV) and LLaVA-Video further broaden this
interface across visual scenarios and larger-scale video instruction tuning
\cite{li2025llavaonevision,zhang2025llavavideo}.
Together, these systems establish the prevailing interface in which encoded
visual features are projected into, and compete for capacity within, the
language-model context.

Scaling the same interface to longer videos makes visual-context management a
central problem. LLaMA-VID uses compact frame representations
\cite{li2024llamavid}, MovieChat consolidates tokens through memory
\cite{song2024moviechat}, and LongVA transfers language-model long-context
capability to the visual modality \cite{zhang2025longcontext}. AVIOT operates
within this VideoLM interface and constructs encoded frame observations into a
compact visual context before decoding.

\subsection{Visual Token Compression in MLLMs}

Visual token compression reduces the context processed by an MLLM either
before or within its language model. Source-side methods select or prune tokens
using attention, language relevance, or visual redundancy
\cite{chen2024fastv,zhang2025sparsevlm,yang2025visionzip}, while PACT and
LLaVA-PruMerge combine removal with clustering or merging
\cite{dhouib2025pact,shang2025llavaprumerge}. Video-specific variants further
exploit temporal redundancy and question relevance. LongVU and Q-Frame combine
temporal reduction with text- or query-guided selection
\cite{shen2025longvu,zhang2025qframe}; PruneVid, FastVID, and FrameFusion use
temporal similarity, segmentation, or importance to prune and merge video tokens
\cite{huang2025prunevid,shen2025fastvid,fu2025framefusion}. Across these
variants, the compact context remains organized around retained source tokens or
source-derived representatives.

Learned bottlenecks offer another route to compact visual tokens.
TokenLearner extracts adaptive token sets \cite{ryoo2021tokenlearner}, while the
Perceiver Resampler and Q-Former map visual features into learned latent sets
\cite{alayrac2022flamingo,li2023blip2}. The Matryoshka Query Transformer and
TokenPacker respectively support variable output cardinality and coarse-to-fine
feature injection \cite{hu2024mqt,li2025tokenpacker}. These outputs need not
retain source-token identities. Recent studies also apply optimal transport to
select representative source subsets or enrich source-derived anchors
\cite{chen2026otprune,li2026aot}. AVIOT instead defines the prescribed compact
representation as the transport target and uses the coupling to construct every
output support.

\begin{figure*}[t]
    \centering
    \includegraphics[width=\textwidth]{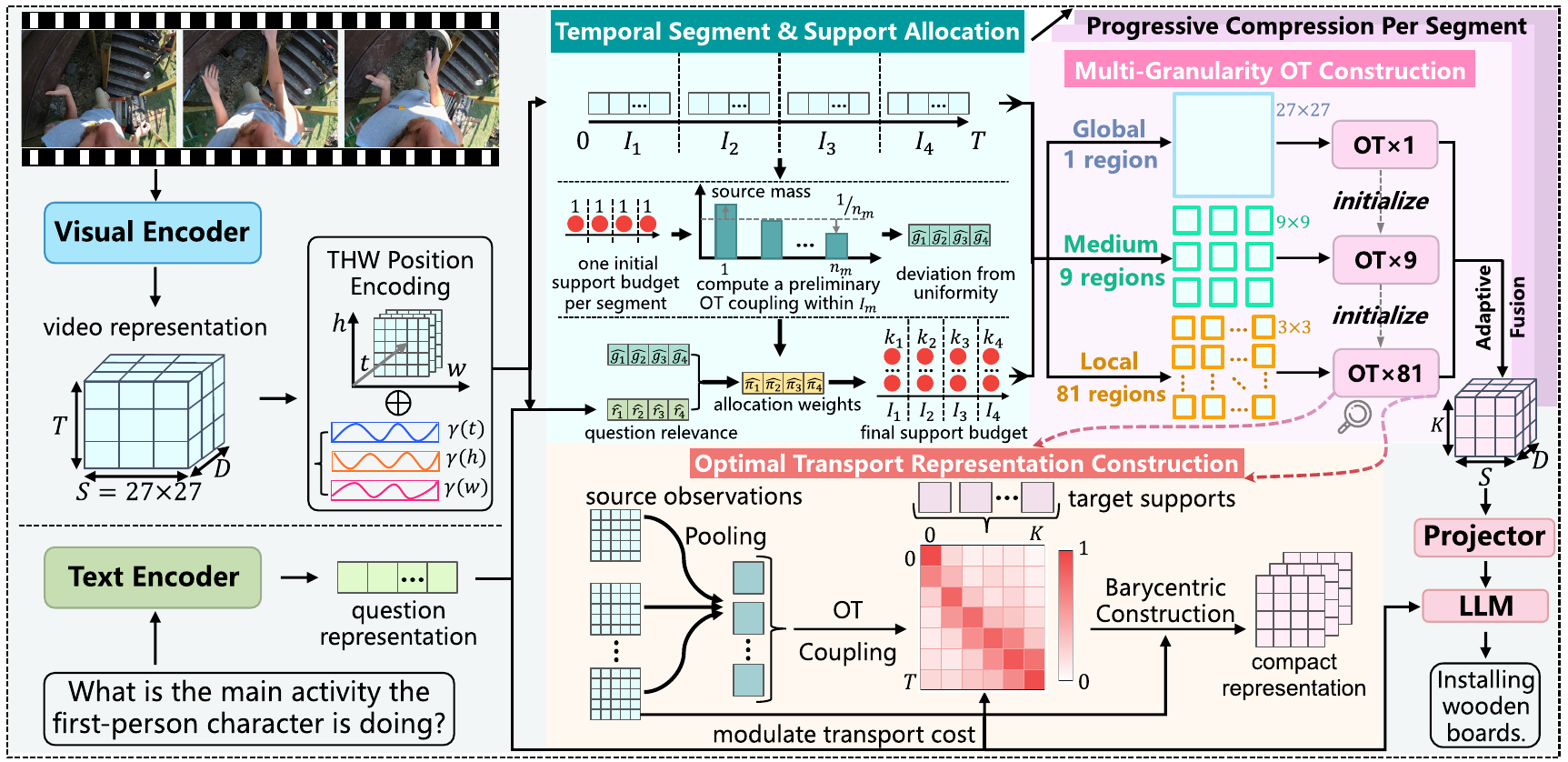}
    \caption{Overview of the proposed AVIOT framework. Dense video features are
    augmented with time--height--width (THW) positional encodings and
    progressively compressed to the requested cardinality. At each stage, the
    question modulates the transport cost and, together with preliminary
    transport responses, guides support allocation across temporal segments.
    Within each segment, source-to-target couplings construct aligned
    representations at multiple spatial granularities through barycentric
    aggregation; these representations are adaptively fused before entering the
    language model.}
    \label{fig:framework}
\end{figure*}

\section{Method}
\label{sec:method}

Video language models answer questions about videos by conditioning
language-model decoding on visual features extracted from sampled frames. For
an input video $\mathcal V$ and textual question $\mathcal Q$, the visual
encoder produces a dense sequence of spatial feature grids
$\mathbf X\in\mathbb R^{T\times S\times D}$, where $T$ indexes sampled frames,
$S$ indexes aligned spatial positions, and $D$ is the visual feature dimension.
Video token compression therefore aims to reduce the visual sequence before
language-model inference.
Conditioned on $\mathcal Q$, AVIOT maps $\mathbf X$ to
$K$ compact visual representations
$\widetilde{\mathbf X}\in\mathbb R^{K\times S\times D}$, where $K<T$.
Each of the $K$ outputs retains the $S$-position spatial layout. Within a given
output, different spatial regions may draw from different source moments.
AVIOT formulates this mapping as aggregating visual
information through optimal transport (OT), viewing dense frame observations
as an empirical source measure and the compact representation as a target
measure with $K$ supports.

AVIOT first augments the feature grids with a time--height--width
(THW) positional encoding and then applies progressive compression toward the
requested cardinality $K$. At each stage, AVIOT divides the current temporal
axis into contiguous segments $\{\mathcal I_m\}_{m=1}^{M}$ and assigns $k_m$
target supports to each segment based on the deviation of its preliminary
transport response from uniformity and the relevance of its content to the
question.
Within every segment, AVIOT estimates a question-conditioned OT coupling and
uses its column-normalized weights to construct the target supports by
barycentric projection. The same OT representation construction is performed
at multiple spatial granularities, and the resulting
representations are adaptively fused to form the compressed representation.
The final representation then follows the multimodal projection and
spatial-pooling path into the language model.
Figure~\ref{fig:framework} summarizes the pipeline.

\subsection{OT Representation Construction}
\label{sec:method_ot}

We first instantiate the core OT representation construction within a temporal segment. Let
$\mathbf X\in\mathbb R^{N\times S\times D}$ denote the spatial features of its
$N$ frames,
and let $k$ be the number of target supports assigned to it. We summarize each
frame observation by spatial pooling as
$\mathbf x_i=\operatorname{Pool}_{s=1}^{S}\mathbf X_{i,s}\in\mathbb R^D$ for
$i=1,\ldots,N$. These descriptors form a uniformly weighted empirical source
measure, while the compact side is represented by a uniformly weighted target
measure,
\begin{equation}
    \mu=\frac{1}{N}\sum_{i=1}^{N}\delta_{\mathbf x_i},
    \qquad
    \nu=\frac{1}{k}\sum_{j=1}^{k}\delta_{\mathbf z_j}.
    \label{eq:source_target_measures}
\end{equation}
whose target supports $\{\mathbf z_j\}_{j=1}^{k}$ lie in the descriptor
space. The supports are initialized at uniformly spaced temporal locations,
denoted by $\mathbf z_{j,\mathrm{uni}}$, and subsequently refined through the
transport coupling. For fixed target supports,
the transport cost between source observation $\mathbf x_i$ and target support
$\mathbf z_j$ is
$C_{ij}(\mathbf q)=d_\theta(\mathbf x_i,\mathbf z_j;\mathbf q)$. The
question-conditioned form of $d_\theta$ is defined in
Sec.~\ref{sec:method_question}. We obtain the source-to-target coupling by
solving the balanced entropy-regularized optimal transport problem,
\begin{equation}
    \mathbf P^\star
    =\underset{\mathbf P\in\Pi}{\arg\min}
    \;\langle\mathbf P,\mathbf C(\mathbf q)\rangle
    +\varepsilon\sum_{i,j}P_{ij}(\log P_{ij}-1),
    \label{eq:entropic_ot}
\end{equation}
where
\begin{equation}
    \Pi
    =\left\{\mathbf P\geq0:\;
    \mathbf P\mathbf 1_k=\frac{1}{N}\mathbf 1_N,\;
    \mathbf P^\top\mathbf 1_N=\frac{1}{k}\mathbf 1_k\right\}.
    \label{eq:transport_polytope}
\end{equation}
The coupling jointly relates all source observations to all target supports.
We approximate $\mathbf P^\star$ with a fixed number of damped log-domain
Sinkhorn updates and denote the resulting coupling by $\widehat{\mathbf P}$.
Numerical solver details are provided in the supplementary material.

Once the coupling has been computed, we normalize the incoming transport mass
of each target support $j$ over the source observations,
\begin{equation}
    A_{ij}
    =\frac{\widehat P_{ij}}
    {\sum_{i'=1}^{N}\widehat P_{i'j}},
    \qquad \sum_{i=1}^{N}A_{ij}=1.
    \label{eq:target_conditional}
\end{equation}
Thus, each column $\mathbf A_{:,j}$ defines the distribution over source
observations associated with one target support. We use this distribution to
refine the support by the barycentric update
$\mathbf z_j\leftarrow\sum_{i=1}^{N}A_{ij}\mathbf x_i$. The updated supports
define the target locations used to recompute the transport cost in the next
round. Starting from their temporal initialization, AVIOT alternates coupling
estimation and barycentric refinement for a small, fixed number of rounds.
After the final round, $\mathbf z_j$ is the refined target support in descriptor
space, while $\mathbf A_{:,j}$ specifies how features from the source frames are
combined at each spatial position to form the compact representation.

Finally, we apply the target-wise distributions to the full spatial features.
For every target support and aligned spatial position, we construct
\begin{equation}
    \mathbf Y_{j,s}
    =\sum_{i=1}^{N}A_{ij}\mathbf X_{i,s},
    \qquad
    \mathbf Y\in\mathbb R^{k\times S\times D}.
    \label{eq:full_grid_construction}
\end{equation}
Consequently, correspondence estimated from frame descriptors produces a
compact representation that retains the full spatial layout. Each
$\mathbf Y_j\in\mathbb R^{S\times D}$ is constructed from the source
observations assigned to its target support.
Equation~\eqref{eq:full_grid_construction} completes the representation
construction for any temporal segment with an assigned number of supports.

\subsection{Question-Conditioned Transport}
\label{sec:method_question}

Within the source-to-target transport construction, question conditioning acts
at two levels. It modulates the transport cost that establishes correspondence
within a temporal segment, while influencing how many target supports are
allocated to each segment. These two paths adapt the composition of the compact
representation to the current question.

\paragraph{Question-modulated transport cost.}
Let $\{\mathbf e_l\}_{l=1}^{L}$ be the language-model embeddings of the
question tokens. We obtain a question representation and project it to the
visual feature dimension,
\begin{equation}
    \bar{\mathbf e}=\frac{1}{L}\sum_{l=1}^{L}\mathbf e_l,
    \qquad
    \mathbf q=\mathbf W_q\bar{\mathbf e}\in\mathbb R^D.
    \label{eq:question_representation}
\end{equation}
A learned map $\phi:\mathbb R^D\rightarrow\mathbb R^{D_c}$ embeds source
descriptors and target supports into a shared transport space. The question
produces positive coordinate-wise weights
\begin{equation}
    \mathbf w(\mathbf q)
    =\operatorname{softplus}(h_\theta(\mathbf q))
    \in\mathbb R_{>0}^{D_c},
    \label{eq:question_weights}
\end{equation}
which specify the question-conditioned transport cost as
\begin{equation}
    C_{ij}(\mathbf q)
    =\left\|
    \mathbf w(\mathbf q)\odot
    \left[\phi(\mathbf x_i)-\phi(\mathbf z_j)\right]
    \right\|_2^2.
    \label{eq:question_transport_cost}
\end{equation}
The question thereby changes the relative contribution of transport-space
dimensions when matching source observations with target supports. Its effect
propagates through the coupling in Eq.~\eqref{eq:entropic_ot} to both support
refinement and representation construction.

\paragraph{Question-conditioned temporal allocation.}
For the temporal segments $\{\mathcal I_m\}_{m=1}^{M}$ introduced above, let
$k_m$ be the number of target supports assigned to segment $m$ at target
cardinality $K$, with $1\leq k_m\leq |\mathcal I_m|$ and
$\sum_{m=1}^{M}k_m=K$. The allocation combines the nonuniformity of each
segment's transport response with its relevance to the question. To allocate
the remaining supports, AVIOT first assigns one target support to each nonempty
segment and computes a preliminary coupling under the same question-conditioned
cost. For each $i\in\mathcal I_m$, summing this coupling over the target
dimension yields the realized source-side row mass, which we normalize within
the segment as
\begin{equation}
    u_i^{(m)}=\sum_j\widehat P_{ij}^{\mathrm{pre},m},
    \qquad
    p_i^{(m)}=\frac{u_i^{(m)}}
    {\sum_{i'\in\mathcal I_m}u_{i'}^{(m)}}.
    \label{eq:pilot_source_profile}
\end{equation}
We measure the deviation of this profile from uniformity as
\begin{equation}
    g_m=\frac{1}{|\mathcal I_m|}
    \sum_{i\in\mathcal I_m}
    \left|p_i^{(m)}-\frac{1}{|\mathcal I_m|}\right|.
    \label{eq:pilot_segment_statistic}
\end{equation}
In parallel, we average frame-level cosine similarities between the question
and projected visual descriptors within each segment to obtain its relevance
score $r_m$. After standardizing both signals, we form
\begin{equation}
    \ell_m=\widehat g_m+\alpha_q\widehat r_m,
    \qquad
    \pi_m=\frac{\exp(\ell_m/\tau)}
    {\sum_{m'=1}^{M}\exp(\ell_{m'}/\tau)}.
    \label{eq:segment_allocation_weights}
\end{equation}
Each nonempty segment retains its initially reserved support, and the remaining
supports are assigned according to $\boldsymbol\pi$. The preliminary response and question
relevance thereby jointly shape the temporal distribution of target supports.

We apply the OT construction of Sec.~\ref{sec:method_ot} independently within
each segment using its assigned $k_m$ supports, and concatenate the resulting
representations in temporal order. The question thus modulates fine-grained
source-to-target correspondence within segments and the allocation of target
supports across segments.

\subsection{Transport across Spatial Granularities}
\label{sec:method_spatial}

Temporal correspondence can depend on the spatial context in which source
observations are compared. The visual representation uses $S$ spatial positions
arranged as a $27\times27$ patch layout. AVIOT instantiates the core OT
construction at three granularities:
global (the full layout), medium (a $3\times3$ partition), and local (a
$9\times9$ partition). Let $\mathcal R^\ell$ be the partition at granularity
$\ell\in\{G,M,L\}$ and let $r_\ell(s)$ denote the region containing spatial
position $s$. For region $r\in\mathcal R^\ell$, we construct temporal source
descriptors $\mathbf x_i^{\ell,r}
=\operatorname{Pool}_{s\in r}\mathbf X_{i,s}$.
Using the segment assignments from Sec.~\ref{sec:method_question}, each region
applies the core OT representation construction of Sec.~\ref{sec:method_ot} to
its descriptors; we use $\widehat{\mathbf P}^{\ell,r}$ and
$\mathbf A^{\ell,r}$ to denote its transport plan and target-wise
distributions, respectively. The global plan is shared across the full layout,
whereas medium and local plans resolve temporal correspondence within their
respective regions after parent-guided support initialization.

Because the spatial partitions form a hierarchy, we transfer coarse temporal
structure to finer transport problems by initializing each child region's target
supports as a combination of parent-induced and uniformly spaced temporal
supports:
\begin{equation}
    \begin{aligned}
        \mathbf z_{j,\mathrm{parent}}^{\ell,r}
        &=\sum_i A_{ij}^{\operatorname{pa}(r)}\mathbf x_i^{\ell,r},\\
        \mathbf z_{j,0}^{\ell,r}
        &=\beta_\ell\mathbf z_{j,\mathrm{parent}}^{\ell,r}
        +(1-\beta_\ell)\mathbf z_{j,\mathrm{uni}}^{\ell,r}.
    \end{aligned}
    \label{eq:coarse_to_fine_initialization}
\end{equation}
We then estimate each child plan from regional descriptors under the
question-modulated transport cost, preserving coarse-to-fine temporal
organization while allowing each spatial context to establish its own
correspondence.

We then apply each region's target-wise distribution to its spatial features
and restore their original order,
\begin{equation}
    \mathbf Y_{j,s}^{\ell}
    =\sum_i A_{ij}^{\ell,r_\ell(s)}\mathbf X_{i,s},
    \qquad
    \mathbf Y^G,\mathbf Y^M,\mathbf Y^L
    \in\mathbb R^{K\times S\times D}.
    \label{eq:granularity_representations}
\end{equation}
The three representations share the same target indices and spatial layout,
but are constructed from temporal plans estimated with different spatial
contexts. A lightweight gate predicts a normalized weight for each granularity at every
target index $j$ and spatial position $s$,
\begin{equation}
    \gamma_{j,s}^{\ell}\geq0,
    \qquad
    \sum_{\ell\in\{G,M,L\}}\gamma_{j,s}^{\ell}=1.
    \label{eq:fusion_weights}
\end{equation}
The gate predicts these weights from compact statistics of the aligned
representations and transport plans, together with the relevance of each
representation to the question. We construct the compressed video
representation as
\begin{equation}
    \widetilde{\mathbf X}_{j,s}
    =\sum_{\ell\in\{G,M,L\}}
    \gamma_{j,s}^{\ell}\mathbf Y_{j,s}^{\ell}.
    \label{eq:granularity_fusion}
\end{equation}
Together, the aligned branches provide complementary global, medium, and local
views of the source observations contributing to each compact
representation.

\subsection{Compression Settings and Training Objective}
\label{sec:method_progressive}

Given a requested compression ratio $r>1$, AVIOT sets the final temporal
cardinality to
$K=\min\{T,\max\{1,\lceil T/r\rceil\}\}$. During training, each example is
assigned a ratio drawn uniformly from $2$ to $10$ in increments of $0.5$, so
that one model is optimized across different output cardinalities. At
inference, the ratio is specified externally and may take any desired value.
AVIOT reaches the resulting $K$ progressively through temporal cardinalities
$T=T_0>T_1>\cdots>T_J=K$, using intermediate cardinalities from $0.75T$,
$0.50T$, and $0.25T$ as applicable. Each stage reapplies the same construction
to the current representation.

Let $\mathcal L_{\mathrm{LM}}$ denote the autoregressive answer-generation loss and
$\mathcal L_{\mathrm{comp}}^{(u)}$ the compression objective at progressive
stage $u$. We optimize
\begin{equation}
    \mathcal L
    =\mathcal L_{\mathrm{LM}}
    +\lambda_{\mathrm{OT}}\frac{1}{J}
    \sum_{u=1}^{J}\mathcal L_{\mathrm{comp}}^{(u)}.
    \label{eq:training_objective}
\end{equation}
The compression objective combines coupling-weighted distortion over global and
regional supports, continuity between neighboring regional plans, smoothness of
spatial fusion weights, and a penalty against granularity collapse. The final
$K\times S\times D$ representation follows the inherited projection and
spatial-pooling path into the language model. Exact solver, objective, and
architecture settings are provided in the experimental setup and supplementary
material.

\section{Experiments}
\label{sec:experiments}

\subsection{Experimental Setup}
\label{sec:experimental_setup}

We evaluate AVIOT on ten video-understanding benchmarks. The controlled
comparison uses Video-MME \cite{fu2025videomme}, EgoSchema
\cite{mangalam2023egoschema}, MVBench \cite{li2024mvbench}, ActivityNet-QA
\cite{yu2019activitynetqa}, and Perception Test
\cite{patraucean2023perceptiontest}. We further assess the same model on
NExT-QA \cite{xiao2021nextqa}, LongVideoBench
\cite{wu2024longvideobench}, Video-MMMU \cite{hu2025videommmu}, LVBench
\cite{wang2025lvbench}, and TempCompass \cite{liu2024tempcompass}. We follow the
official evaluation protocol of each benchmark and report its aggregate score.
Unless otherwise stated, videos are sampled at 2 FPS with at most 224 frames,
matching the sampling used for training. ActivityNet-QA is evaluated using
GPT-4o as a semantic judge. All values are percentages, and higher is better.

AVIOT is initialized from LLaVA-Video-7B-Qwen2, with
SigLIP-SO400M-patch14-384 as the visual encoder, and trained on
LLaVA-Video-178K for 10,903 steps. We jointly optimize the language model,
multimodal projector, and AVIOT compressor using a global batch size of 128 on
32 NVIDIA H800 GPUs. Training uses BF16, an initial learning rate of
$10^{-5}$, zero weight decay, a cosine schedule, and a warmup ratio of 0.03. We
use four temporal segments and 20 log-domain Sinkhorn updates for each
coupling. Additional architecture and optimization settings are provided in
the supplementary material.

\begin{table}[!ht]
    \centering
    \begingroup
    \small
    \setlength{\tabcolsep}{0pt}
    \def\retcell#1{\makebox[28.5pt][c]{#1}}
    \def\flopcell#1{\makebox[26pt][c]{#1}}
    \def\scorecell#1{\makebox[28pt][c]{#1}}
    \def\stackcell#1{\begin{tabular}[c]{@{}c@{}}#1\end{tabular}}
    \def\copecell#1#2{\scorecell{\stackcell{#1\\#2}}}
    \begin{tabular*}{\columnwidth}{@{\extracolsep{\fill}}c|cc|ccccc@{}}
        \toprule
        \stackcell{Method} &
        \retcell{\stackcell{Token\\ret.}} &
        \flopcell{\stackcell{Prefill\\FLOPs\\(T)}} &
        \scorecell{\stackcell{Acti-\\vity\\Net-\\QA}} &
        \scorecell{\stackcell{Per-\\cep-\\tion\\Test}} &
        \scorecell{\stackcell{MV\\Bench}} &
        \scorecell{\stackcell{Video-\\MME\\w/o\\sub.}} &
        \scorecell{\stackcell{Ego\\Sche-\\ma}} \\
        \midrule
        \stackcell{LLaVA-\\Video-7B} &
        \retcell{100.0\%} & \flopcell{106.69} &
        \scorecell{64.10} & \scorecell{67.90} & \scorecell{58.60} &
        \scorecell{\textbf{63.30}} & \scorecell{57.30} \\
        \stackcell{CoPE-\\VideoLM} &
        \retcell{\stackcell{varying}} & \flopcell{\stackcell{--}} &
        \copecell{64.80}{92\%} &
        \copecell{70.30}{20\%} &
        \copecell{61.90}{16\%} &
        \copecell{61.90}{289\%} & \scorecell{\stackcell{--}} \\
        \midrule
        \multirow{3}{*}{\stackcell{Uniform\\Keep}} & \retcell{50.0\%} & \flopcell{44.28} & \scorecell{67.01} & \scorecell{68.34} & \scorecell{60.75} & \scorecell{62.11} & \scorecell{57.21} \\
        & \retcell{25.0\%} & \flopcell{19.87} & \scorecell{65.79} & \scorecell{67.01} & \scorecell{58.40} & \scorecell{60.30} & \scorecell{54.98} \\
        & \retcell{10.9\%} & \flopcell{8.14} & \scorecell{63.42} & \scorecell{64.46} & \scorecell{55.12} & \scorecell{56.26} & \scorecell{51.38} \\
        \cmidrule(lr){1-8}
        \multirow{3}{*}{\stackcell{Segment\\Mean}} & \retcell{50.0\%} & \flopcell{44.28} & \scorecell{66.40} & \scorecell{67.74} & \scorecell{60.48} & \scorecell{61.22} & \scorecell{55.34} \\
        & \retcell{25.0\%} & \flopcell{19.87} & \scorecell{61.90} & \scorecell{65.24} & \scorecell{58.15} & \scorecell{54.30} & \scorecell{48.22} \\
        & \retcell{10.9\%} & \flopcell{8.14} & \scorecell{53.44} & \scorecell{61.77} & \scorecell{53.60} & \scorecell{48.22} & \scorecell{39.87} \\
        \midrule
        \multirow{3}{*}{AVIOT} & \retcell{50.0\%} & \flopcell{44.28} & \scorecell{\textbf{68.22}} & \scorecell{\textbf{71.50}} & \scorecell{\textbf{62.20}} & \scorecell{63.22} & \scorecell{\textbf{57.86}} \\
        & \retcell{25.0\%} & \flopcell{19.87} & \scorecell{67.06} & \scorecell{70.72} & \scorecell{61.33} & \scorecell{60.85} & \scorecell{56.33} \\
        & \retcell{10.9\%} & \flopcell{8.14} & \scorecell{65.67} & \scorecell{69.41} & \scorecell{58.65} & \scorecell{58.26} & \scorecell{53.61} \\
        \bottomrule
    \end{tabular*}
    \endgroup
    \caption{Controlled comparison with a 64-frame input at compression ratios
    2, 4, and 10. The second column reports token retention; bold denotes the
    best score in each benchmark column.}
    \label{tab:fixed64_results}
\end{table}

\begin{table*}[t]
    \centering
    \begingroup
    \small
    \setlength{\tabcolsep}{1.4pt}
    \begin{tabular*}{\textwidth}{@{\extracolsep{\fill}}lcccccccccc@{}}
        \toprule
        Model & \shortstack{Video-\\MME\\w/o sub.} & \shortstack{Video-\\MMMU} &
        \shortstack{ActivityNet-\\QA} & NExT-QA & \shortstack{Perception\\Test} &
        \shortstack{Temp\\Compass} & LongVideoBench & LVBench & \shortstack{Ego\\Schema} & MVBench \\
        \midrule
        LongVA \shortcite{zhang2025longcontext} & 52.6 & 23.9 & 50.0 & 68.3 & -- & 56.9 & -- & -- & -- & -- \\
        IXC-2.5 \shortcite{zhang2024ixc25} & 55.8 & -- & 52.8 & 71.0 & 34.4 & 67.1 & -- & -- & -- & \textbf{69.1} \\
        LLaVA-OV \shortcite{li2025llavaonevision} & 58.2 & 33.9 & 56.6 & 79.4 & 57.1 & 64.8 & 56.4 & 38.1 & \textbf{60.1} & 56.7 \\
        Apollo \shortcite{zohar2024apollo} & 61.3 & -- & -- & -- & 67.3 & 64.9 & \textbf{58.5} & -- & -- & -- \\
        Oryx \shortcite{liu2025oryx} & 58.3 & -- & -- & 81.9 & 68.6 & -- & 55.3 & -- & -- & \underline{63.9} \\
        \midrule
        LLaVA-Video \shortcite{zhang2025llavavideo} & \underline{63.3} & 36.1 & 64.1 & \underline{83.2} & 67.9 & 66.6 & \underline{58.2} & 44.2 & 57.3 & 58.6 \\
        CoPE \shortcite{sarkar2026copevideolm} & 61.9 & \underline{38.2} & \underline{64.8} & 82.1 & \underline{70.3} & \textbf{68.9} & 56.9 & \textbf{46.4} & -- & 61.9 \\
        \midrule
        AVIOT & \textbf{63.89} & \textbf{39.78} & \textbf{68.23} & \textbf{83.49} & \textbf{71.45} & \underline{68.35} & 56.99 & \underline{45.38} & \underline{58.14} & 61.58 \\
        \bottomrule
    \end{tabular*}
    \endgroup
    \caption{Open-source 7B VideoLM comparison. IXC-2.5 is
    InternLM-XComposer-2.5; bold/underline: best/second best.}
    \label{tab:dense_results}
\end{table*}

\subsection{Main Results}
\label{sec:main_results}

For the controlled comparison in Table~\ref{tab:fixed64_results}, we uniformly
sample 64 frames to match the input setting of the uncompressed
LLaVA-Video-7B backbone \cite{zhang2025llavavideo}. We compare AVIOT with this
baseline and CoPE-VideoLM \cite{sarkar2026copevideolm}, which uses the same
backbone and training corpus and follows its published codec-based sampling
with a dataset-dependent token budget. We also construct two baselines on this
backbone: \emph{Uniform Keep} retains $K$ uniformly spaced source frames,
whereas \emph{Segment Mean} divides the temporal axis into $K$ contiguous
intervals and averages the aligned spatial features within each interval. The
three AVIOT operating points correspond to compression ratios 2, 4, and 10. At
ratios 2 and 4, AVIOT exceeds the uncompressed backbone by 2.36 and 1.02
points on average, respectively. At ratio 2, it improves four of the five
benchmarks and remains within 0.08 points on Video-MME. Even at ratio 10, the
mean decrease is only 1.12 points, while ActivityNet-QA, Perception Test, and
MVBench match or exceed the backbone. AVIOT also outperforms both controlled
baselines on every benchmark at every ratio; at ratio 10, its gains range from
2.00 to 4.95 points over \emph{Uniform Keep} and from 5.05 to 13.74 points over
\emph{Segment Mean}. At a comparable retention on Perception Test, AVIOT
reaches 70.72 at 25.0\% versus CoPE's 70.30 at 20.19\%.

Table~\ref{tab:dense_results} shows that, at $r=4$, AVIOT maintains strong or
competitive performance across benchmarks spanning diverse video types and
understanding tasks. This is achieved using only LLaVA-Video-178K, a
considerably smaller training corpus than those used by most compared
general-purpose VideoLMs. The limited coverage of long videos in this corpus
may partly explain the remaining gaps on LongVideoBench and LVBench.

\subsection{Ablation Studies}
\label{sec:ablation_studies}

On Video-MME at $r=2$, Table~\ref{tab:component_ablation} ablates question
conditioning and transport across spatial granularities at inference time on
the same checkpoint.
Removing the question-conditioned transport cost lowers 65.11 to 64.85, while
removing temporal allocation yields 64.59, showing that both adaptive
correspondence and temporal capacity allocation benefit AVIOT. Removing both
paths retains 64.67, only 0.44 points below AVIOT. This non-additive change
suggests interaction between the two paths, while the retained score shows that
OT construction remains robust when correspondence and support allocation are
question-independent. For spatial granularity, global transport scores 64.77;
regional transport with equal fusion improves it to 64.81, and adaptive fusion
reaches 65.11. The 0.34-point overall gain supports both region-specific
temporal correspondence and learned, rather than equal, fusion across target
supports and spatial positions.

\begin{table}[!ht]
    \centering
    \begingroup
    \small
    \setlength{\tabcolsep}{4pt}
    \begin{tabular*}{\columnwidth}{@{\extracolsep{\fill}}lc@{}}
        \toprule
        Configuration & Score \\
        \midrule
        \multicolumn{2}{l}{\emph{Question conditioning}} \\
        Full AVIOT & 65.11 \\
        w/o question-conditioned transport cost & 64.85 \\
        w/o question-conditioned temporal allocation & 64.59 \\
        w/o both conditioning paths & 64.67 \\
        \midrule
        \multicolumn{2}{l}{\emph{Transport across spatial granularities}} \\
        Global transport only & 64.77 \\
        Multiple spatial granularities, equal fusion & 64.81 \\
        Multiple spatial granularities, adaptive fusion (default) & 65.11 \\
        \bottomrule
    \end{tabular*}
    \endgroup
    \caption{Component analysis of AVIOT.}
    \label{tab:component_ablation}
\end{table}

\subsection{Further Analysis}
\label{sec:further_analysis}

\paragraph{Generalization across compression ratios.}
\begin{figure}[t]
    \centering
    \includegraphics[width=\columnwidth]{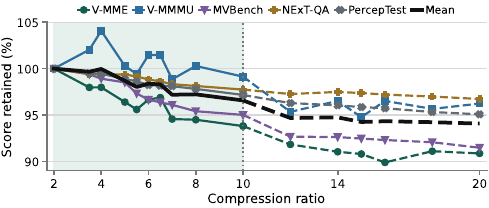}
    \caption{Generalization across compression ratios.}
    \label{fig:ratio_generalization}
\end{figure}

AVIOT is trained with ratios between $2$ and $10$, but its target cardinality
can be specified directly at inference. Figure~\ref{fig:ratio_generalization}
examines ratios from $2$ to $20$ on five benchmarks. The mean score across these
benchmarks changes from 64.38 at $r=2$ to 62.27 at $r=10$ and 60.74 at $r=20$;
doubling the compression beyond the largest training ratio therefore reduces
the mean by 1.53 points. The gradual degradation beyond the training interval
shows that the same checkpoint generalizes to substantially higher, unseen
compression ratios without additional optimization.

\begin{figure}[!b]
    \centering
    \includegraphics[width=\columnwidth]{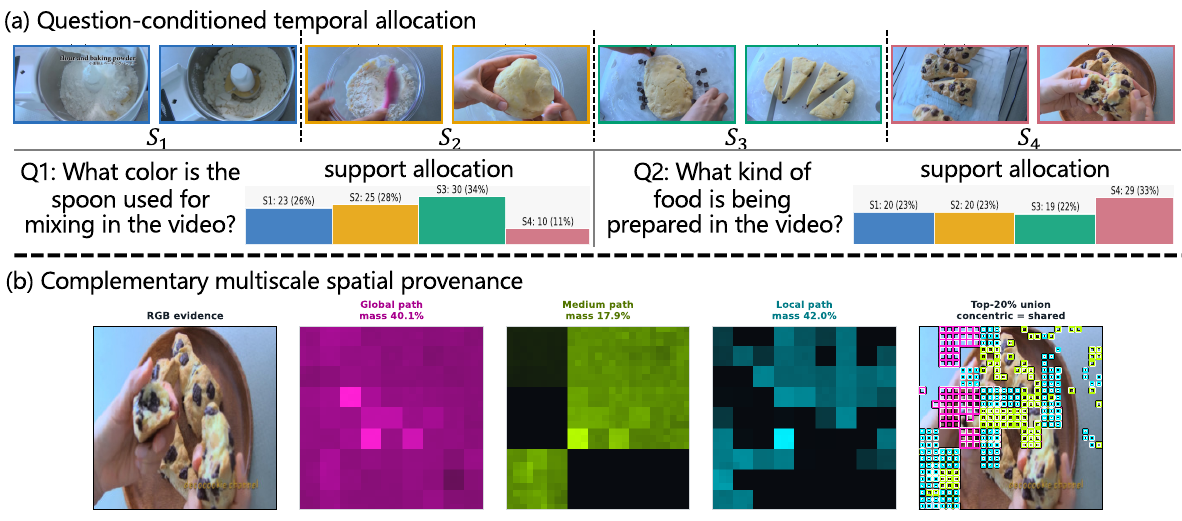}
    \caption{Question-conditioned temporal allocation and provenance across
    spatial granularities. (a) Support allocations for two questions. (b)
    Global, medium, and local provenance for one source frame and target
    support; percentages denote fused mass, and boxes mark the top 20\%
    positions.}
    \label{fig:question_multiscale_provenance}
\end{figure}

\paragraph{Question conditioning and spatial granularity.}
Figure~\ref{fig:question_multiscale_provenance}(a) isolates the effect of
question-conditioned temporal allocation by holding the video, model,
compression ratio, temporal segmentation, and total support budget fixed. For
the question about the spoon's color, the allocation $[23,25,30,10]$ assigns
88.6\% of the supports to the first three temporal segments and 11.4\% to the
final segment. Changing only the question to the type of food being prepared
produces $[20,20,19,29]$, increasing the allocation to the final,
food-revealing segment to 33.0\%. The redistribution shows how question
conditioning directs representation capacity toward different temporal stages
while the overall compression budget remains unchanged.

For the food question, Figure~\ref{fig:question_multiscale_provenance}(b)
traces the contribution from source frame $i=113$ ($56.5\,\mathrm{s}$) to target
support $j=22$ along the three granularity-specific transport paths. After
adaptive fusion, the global, medium, and local paths account for 40.1\%, 17.9\%,
and 42.0\% of the exact provenance mass for this pair, confirming that all three
remain active. Independent heatmap normalization reveals each path's internal
spatial pattern, while the percentages retain their relative cross-path mass.
In the top-20\% union overlay, single-colored boxes mark path-specific positions
and concentric boxes mark agreement across paths. The maps emphasize different
parts of the food, hand, and surrounding context, showing how the aligned paths
combine shared and complementary spatial content within the same target
support.

\paragraph{Hierarchical evidence provenance.}
\begin{figure}[!t]
    \centering
    \includegraphics[width=\columnwidth]{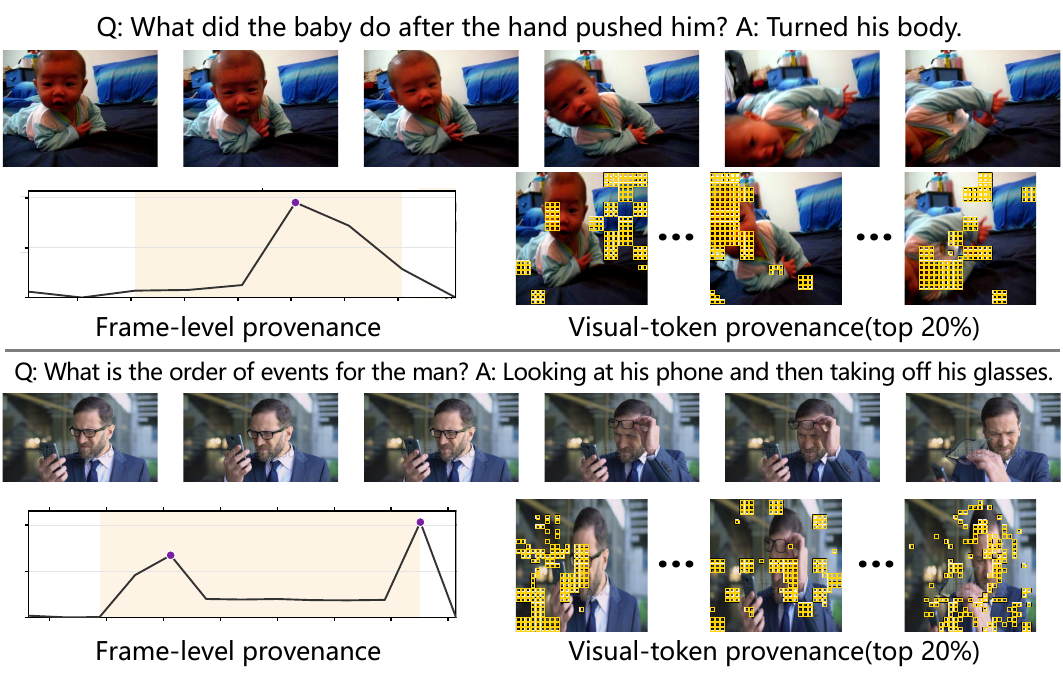}
    \caption{Hierarchical source-frame and spatial provenance of selected target
    supports.
    }
    \label{fig:hierarchical_provenance}
\end{figure}

Figure~\ref{fig:hierarchical_provenance} traces a final target support to its
contributing source frames and spatial positions. For support $j$, its feature
at position $s$ is
$\widetilde{\mathbf X}_{j,s}=\sum_i c^{\mathrm{fused}}_{j,s,i}\mathbf X_{i,s}$,
where $c^{\mathrm{fused}}_{j,s,i}$ is the exact source-frame contribution
obtained by composing the target-wise distributions and adaptive fusion weights
across progressive compression stages. Averaging over spatial positions yields
the frame-level provenance
$\omega_i=S^{-1}\sum_s c^{\mathrm{fused}}_{j,s,i}$. The unaveraged coefficients
form a token-level provenance map for each source frame. The figure presents
the frame- and token-level provenance together: curves plot $\omega_i$ over
source frames, shaded intervals
locate the displayed sequences, and yellow boxes mark positions with top-20\%
coefficients. In the first example, the curve peaks at frame 86
($43.04\,\mathrm{s}$), as the baby begins the answer-defining body turn. Across
the sequence, high-provenance positions follow the face, torso, and moving body,
localizing the evidence as the event unfolds. In the second example, target
support $j=1$ exhibits two separated peaks: frame 5 ($2.50\,\mathrm{s}$)
corresponds to looking at the phone, whereas frame 12 ($6.01\,\mathrm{s}$)
captures the subsequent removal of the glasses. Their token-level maps localize
each contribution, showing that one support integrates disjoint temporal windows
needed to resolve the event order.

\section{Conclusion}

We presented AVIOT, an optimal-transport framework that constructs compact
video representations through a joint source-to-target coupling. Question
conditioning adapts temporal correspondence and support allocation, while
spatial-granularity transport enables region-specific aggregation. Across ten
benchmarks, AVIOT maintains strong performance across compression ratios,
supporting OT-based representation construction for VideoLM token compression.

\bibliography{references}

\end{document}


\pagestyle{empty}
\thispagestyle{empty}
\makesupplementtitle

\section{Supplementary Overview}
\label{sec:supp_overview}

This supplementary material provides additional method details and evidence
for AVIOT. It expands the representation construction and its numerical
realization, specifies the architecture and training objective, and presents
extended quantitative, qualitative, and visual-token accounting analyses.

\section{Additional Method Details}
\label{sec:supp_method}

This section expands the representation construction summarized in the main
paper. We use $\mathbf U\in\mathbb R^{T\times S\times D}$ for the visual
encoder output before positional augmentation and
$\mathbf X\in\mathbb R^{T\times S\times D}$ for the features presented to
AVIOT. The final configuration has $S=27^2$ and $D=1152$. A target support is a
location in descriptor space; its corresponding element of the compressed
video representation retains all $S$ spatial positions.

\subsection{Position-Aware Video Features}
\label{sec:supp_thw}

AVIOT augments every visual feature with explicit time--height--width (THW)
coordinates before computing frame or regional descriptors. For frame $i$ and
spatial position $s$, let $t_i,h_s,w_s\in[-1,1]$ be coordinates sampled
uniformly along their respective axes. We scale the temporal coordinate by
$a_t=1$ and both spatial coordinates by $a_h=a_w=0.5$. For a scalar coordinate
$c$, the Fourier map with $B=16$ bands is
\begin{equation}
    \psi(c)=
    \bigl[\sin(2^b\pi c),\cos(2^b\pi c)\bigr]_{b=0}^{B-1}.
    \label{eq:supp_fourier_map}
\end{equation}
Concatenating the three maps gives a 96-dimensional positional vector. A
learned bias-free projection maps it to the visual feature dimension, and the
position-aware feature is
\begin{equation}
    \mathbf X_{i,s}=\mathbf U_{i,s}
    +\alpha_{\mathrm{pos}}\mathbf W_{\mathrm{pos}}
    [\psi(a_t t_i);\psi(a_h h_s);\psi(a_w w_s)].
    \label{eq:supp_thw_residual}
\end{equation}
The projection is initialized to zero. During the first 500 optimization steps,
$\alpha_{\mathrm{pos}}$ increases linearly from zero to one; it equals one
thereafter and at inference. Thus, the positional term is introduced as a
residual without changing the visual encoder or the shape of its output.

\subsection{Question Conditioning and Temporal Allocation}
\label{sec:supp_question_allocation}

Question conditioning uses the question-token span, excluding visual
placeholders and supervised answer tokens. If
$\{\mathbf e_l\}_{l=1}^{L}$ are its language-model embeddings, their mean is
detached from the embedding lookup and projected to the visual dimension:
\begin{equation}
    \bar{\mathbf e}=L^{-1}\sum_{l=1}^{L}\mathbf e_l,
    \qquad
    \mathbf q_v=\mathbf W_q\bar{\mathbf e}\in\mathbb R^D.
    \label{eq:supp_question_vectors}
\end{equation}
The learned projection remains trainable. The transport metric maps source and
target descriptors to $D_c=256$ dimensions with a two-layer MLP $\phi$. A
second MLP maps $\mathbf q_v$ to positive coordinate weights, giving
\begin{equation}
\begin{aligned}
    \mathbf w_q&=\operatorname{softplus}(h_\theta(\mathbf q_v)),\\
    C_{ij}(\mathbf q_v)&=
    \left\|\mathbf w_q\odot
    [\phi(\mathbf x_i)-\phi(\mathbf z_j)]\right\|_2^2.
\end{aligned}
    \label{eq:supp_question_cost}
\end{equation}
The same $\mathbf q_v$ conditions all spatial granularities in a compression
stage.

At a stage with $N$ source observations and target cardinality $K$, the
temporal axis is divided into
$M=\min\{4,N,K\}$ contiguous, nearly equal segments. Earlier segments receive
one additional observation when $N$ is not divisible by $M$. Each segment is
first assigned one target support, so its preliminary coupling has one target
column. For segment $m$, let $\widehat{\mathbf P}^{\mathrm{pre},m}$ denote this
coupling. The realized row mass and its within-segment normalization are
\begin{equation}
    u_i^{(m)}=\sum_j\widehat P_{ij}^{\mathrm{pre},m},
    \qquad
    p_i^{(m)}=\frac{u_i^{(m)}}
    {\sum_{i'\in\mathcal I_m}u_{i'}^{(m)}}.
    \label{eq:supp_pilot_profile}
\end{equation}
The pilot response statistic is the mean absolute departure from a uniform
profile,
\begin{equation}
    g_m=\frac{1}{|\mathcal I_m|}
    \sum_{i\in\mathcal I_m}
    \left|p_i^{(m)}-\frac{1}{|\mathcal I_m|}\right|.
    \label{eq:supp_pilot_deviation}
\end{equation}

The second allocation signal is computed in the language-model feature space.
Each position-aware frame is passed through the multimodal projector and
spatially averaged to obtain $\mathbf v_i$. With the unprojected question mean
$\bar{\mathbf e}$, frame relevance and segment relevance are
\begin{equation}
    \rho_i=\cos(\mathbf v_i,\bar{\mathbf e}),
    \qquad
    r_m=\frac{1}{|\mathcal I_m|}
    \sum_{i\in\mathcal I_m}\rho_i.
    \label{eq:supp_segment_relevance}
\end{equation}
This relevance path is used only to inform temporal support allocation; the
transport cost itself is conditioned by Eq.~\eqref{eq:supp_question_cost}.

Let $\mathcal N(\cdot)$ standardize a vector over the current segments by
subtracting its mean and dividing by its population standard deviation. If the
standard deviation is numerically zero, the implementation uses the maximum
absolute centered value, and returns zero when no segment differs. The
allocation probabilities are
\begin{equation}
\begin{aligned}
    \ell_m&=\mathcal N(\mathbf g)_m
    +\alpha_q\mathcal N(\mathbf r)_m,\\
    \pi_m&=\frac{\exp(\ell_m/\tau)}
    {\sum_{m'=1}^{M}\exp(\ell_{m'}/\tau)}.
\end{aligned}
    \label{eq:supp_allocation_probability}
\end{equation}
During the first 500 steps, $\alpha_q$ increases linearly from zero to $0.3$
and $\tau$ decreases linearly from $1.0$ to $0.1$. The remaining $K-M$
supports are apportioned by rounding $\pi_m(K-M)$, capped by the number of
available observations in each segment. A deterministic repair then adds
supports in descending $\pi_m$ order, or removes them in ascending order, until
\begin{equation}
    1\leq k_m\leq|\mathcal I_m|,
    \qquad
    \sum_{m=1}^{M}k_m=K.
    \label{eq:supp_integer_budget}
\end{equation}
All segment outputs are concatenated in temporal order.

\subsection{Transport across Spatial Granularities}
\label{sec:supp_spatial}

The three spatial partitions contain one $27\times27$ global region, nine
$9\times9$ medium regions, and 81 $3\times3$ local regions. For a region
$r\in\mathcal R^\ell$, its descriptor at source time $i$ is
\begin{equation}
    \mathbf x_i^{\ell,r}=
    \operatorname{LN}\!\left(
    \frac{1}{|r|}\sum_{s\in r}\mathbf X_{i,s}\right)
    +\mathbf p_r.
    \label{eq:supp_region_descriptor}
\end{equation}
Here $\mathbf p_r$ is a learned projection of sine and cosine encodings of the
region-center coordinates. It uses integer frequencies $1,\ldots,8$ along
each spatial axis and is initialized to zero. The global descriptor is the
spatial mean used by the core construction.

Medium supports combine support locations induced by the global
target-conditional distributions with uniform temporal initialization; local
supports are initialized analogously from their medium parent. For child
granularity $\ell$ and region $r$,
\begin{equation}
    \begin{aligned}
    \mathbf z_{j,\mathrm{par}}^{\ell,r}
      &=\sum_i A_{ij}^{\operatorname{pa}(r)}
        \mathbf x_i^{\ell,r},\\
    \mathbf z_{j,0}^{\ell,r}
      &=\beta_\ell\mathbf z_{j,\mathrm{par}}^{\ell,r}
        +(1-\beta_\ell)\mathbf z_{j,\mathrm{uni}}^{\ell,r},
    \end{aligned}
    \label{eq:supp_parent_initialization}
\end{equation}
with $\beta_M=0.50$ and $\beta_L=0.25$. Regional transport uses the same
question-conditioned metric as Eq.~\eqref{eq:supp_question_cost}, with
$\varepsilon_M=0.12$ and $\varepsilon_L=0.15$. All nine medium and all 81
local regions are evaluated. Their target-conditional distributions reconstruct
aligned branch representations
\begin{equation}
    \mathbf Y_{j,s}^{\ell}
    =\sum_i A_{ij}^{\ell,r_\ell(s)}\mathbf X_{i,s},
    \qquad \ell\in\{G,M,L\}.
    \label{eq:supp_spatial_reconstruction}
\end{equation}

The adaptive gate is evaluated independently for every target support $j$ and
spatial position $s$. For each branch it uses seven compact statistics: relative
change from the adjacent target support, normalized transport entropy, relative
RMS distance and cosine distance from the three-branch mean, cosine similarity
to $\mathbf q_v$, relative log feature magnitude, and a fixed granularity
identity. Concatenating these statistics for the three branches yields a
21-dimensional vector. After layer normalization, a two-layer MLP with hidden
dimension 32 predicts three logits. With temperature $\tau_g=0.05$, let
$\mathbf p_{j,s}=\operatorname{softmax}(\mathbf a_{j,s}/\tau_g)$. A global
floor $\delta_G=0.20$ gives
\begin{equation}
    \gamma_{j,s}^{G}=\delta_G+(1-\delta_G)p_{j,s}^{G},
    \quad
    \gamma_{j,s}^{M,L}=(1-\delta_G)p_{j,s}^{M,L}.
    \label{eq:supp_gate_floor}
\end{equation}
The weights are nonnegative and sum to one. The stage output is therefore
$\widetilde{\mathbf X}_{j,s}=
\sum_{\ell}\gamma_{j,s}^{\ell}\mathbf Y_{j,s}^{\ell}$. During the first 500
training steps, the regional contribution is linearly introduced by
interpolating this output with the global branch; the full fused output is used
thereafter and at inference.

\subsection{Progressive Construction and Exact Provenance}
\label{sec:supp_progressive}

For requested ratio $r$, AVIOT uses
$K=\min\{T,\max\{1,\lceil T/r\rceil\}\}$. Candidate intermediate
cardinalities are $0.75T$, $0.50T$, and $0.25T$, rounded to the nearest multiple
of eight and retained only when they lie strictly between the current
cardinality and $K$. The final $K$ is always appended. For $T=224$, the verified
paths are
\begin{equation}
\begin{array}{ccl}
r=2  &:&224\rightarrow168\rightarrow112,\\
r=4  &:&224\rightarrow168\rightarrow112\rightarrow56,\\
r=10 &:&224\rightarrow168\rightarrow112\rightarrow56\rightarrow23,\\
r=20 &:&224\rightarrow168\rightarrow112\rightarrow56\rightarrow12.
\end{array}
\label{eq:supp_verified_schedules}
\end{equation}

The output of one stage becomes the source representation of the next. To make
the resulting provenance explicit, let
$B^{(u)}_{j,s,i}$ denote the feature-mixture coefficient from source $i$ to
target $j$ at spatial position $s$ in stage $u$. It combines the
granularity-specific target-conditional distributions and fusion weights:
\begin{equation}
    B^{(u)}_{j,s,i}=
    \sum_{\ell\in\{G,M,L\}}
    \gamma_{j,s}^{(u),\ell}
    A_{ij}^{(u),\ell,r_\ell(s)}.
    \label{eq:supp_stage_provenance}
\end{equation}
Each coefficient is nonnegative and
$\sum_iB^{(u)}_{j,s,i}=1$. Starting with
$C^{(1)}=B^{(1)}$, coefficients relative to the original source video compose
recursively as
\begin{equation}
    C^{(u)}_{j,s,i}=
    \sum_h B^{(u)}_{j,s,h}C^{(u-1)}_{h,s,i}.
    \label{eq:supp_progressive_composition}
\end{equation}
Consequently, the final feature satisfies
\begin{equation}
    \widetilde{\mathbf X}_{j,s}
    =\sum_{i=1}^{T}C^{(J)}_{j,s,i}\mathbf X_{i,s},
    \qquad \sum_iC^{(J)}_{j,s,i}=1.
    \label{eq:supp_exact_provenance}
\end{equation}
These are exact feature-mixture coefficients produced by the executed
construction, rather than post-hoc attention or saliency estimates.

\section{Numerical OT Construction}
\label{sec:supp_numerical_ot}

The main paper states the entropy-regularized OT objective and its nominal
uniform source and target marginals. We approximate its coupling using
finite-step log-domain Sinkhorn updates~\citep{cuturi2013sinkhorn}. This section
specifies these updates and the explicit normalization used to construct each
target support.

\subsection{Finite-Step Log-Domain Updates}
\label{sec:supp_log_updates}

For a segment containing $N$ source observations and $k$ target supports, let
$a_i=1/N$ and $b_j=1/k$. Before the log-domain recurrence, the cost is bounded
relative to the entropy scale:
\begin{equation}
    \widetilde{\mathbf C}=s_C\mathbf C,
    \qquad
    s_C=\min\left\{1,
    \frac{100\varepsilon}{\max_{i,j}C_{ij}+10^{-8}}\right\}.
    \label{eq:supp_cost_stabilization}
\end{equation}
This common rescaling is active only when the largest cost exceeds
$100\varepsilon$. Define the two log-domain marginal maps
\begin{align}
    \mathcal T_a(\mathbf g)_i
    &=\varepsilon\log a_i-\varepsilon
      \log\sum_j\exp\!\left(
      \frac{g_j-\widetilde C_{ij}}{\varepsilon}\right),
      \label{eq:supp_row_map}\\
    \mathcal T_b(\mathbf f)_j
    &=\varepsilon\log b_j-\varepsilon
      \log\sum_i\exp\!\left(
      \frac{f_i-\widetilde C_{ij}}{\varepsilon}\right).
      \label{eq:supp_col_map}
\end{align}
Starting from $\mathbf f^{(0)}=\mathbf 0$ and
$\mathbf g^{(0)}=\mathbf 0$, AVIOT performs
\begin{equation}
\begin{aligned}
    \mathbf f^{(n+1)}
      &=(1-\kappa_s)\mathbf f^{(n)}
        +\kappa_s\mathcal T_a(\mathbf g^{(n)}),\\
    \mathbf g^{(n+1)}
      &=(1-\kappa_t)\mathbf g^{(n)}
        +\kappa_t\mathcal T_b(\mathbf f^{(n+1)}),
\end{aligned}
\label{eq:supp_damped_updates}
\end{equation}
where
\begin{equation}
    \kappa_s=\frac{\rho_s}{\rho_s+\varepsilon},
    \qquad
    \kappa_t=\frac{\rho_t}{\rho_t+\varepsilon}.
    \label{eq:supp_damping_coefficients}
\end{equation}
The final configuration uses $\rho_s=0.5$, $\rho_t=5.0$, and 20 updates. For
the global entropy coefficient $\varepsilon=0.10$, this gives
$\kappa_s=5/6$ and $\kappa_t=50/51$. The finite-step coupling is
\begin{equation}
    \widehat P_{ij}=
    \exp\!\left(
    \frac{f_i^{(20)}+g_j^{(20)}-\widetilde C_{ij}}{\varepsilon}
    \right).
    \label{eq:supp_finite_coupling}
\end{equation}
The damped updates approach the same marginal maps as the standard fixed-point
iteration. AVIOT uses the resulting finite-step coupling directly, and the
column normalization below converts each realized target column into the
target-conditional distribution used for representation construction.

\subsection{Target-Conditional Distributions and Refinement}
\label{sec:supp_reconstruction}

Each target column is normalized explicitly:
\begin{equation}
    A_{ij}=\frac{\widehat P_{ij}}
    {m_j},
    \qquad
    m_j=\sum_{i'=1}^{N}\widehat P_{i'j}.
    \label{eq:supp_column_normalization}
\end{equation}
The strictly positive exponential coupling gives $m_j>0$, and hence
$\sum_iA_{ij}=1$. With the numerical stabilizer $\delta=10^{-8}$, support
refinement and the final spatial representation are constructed as
\begin{equation}
    \mathbf z_j\leftarrow
    \frac{\sum_i\widehat P_{ij}\mathbf x_i}{m_j+\delta},
    \qquad
    \mathbf Y_{j,s}=\sum_iA_{ij}\mathbf X_{i,s}.
    \label{eq:supp_barycentric_reconstruction}
\end{equation}
AVIOT recomputes the question-conditioned cost after each support update. The
core global construction uses five refinement rounds. The regional hierarchy
receives this global plan and performs two refinement rounds for each medium
region and one for each local region. The corresponding entropy coefficients
are $0.10$, $0.12$, and $0.15$ for global, medium, and local transport.

When a segment already has as many target supports as source observations, the
stage uses the identity plan. This preserves the segment exactly and avoids an
unnecessary numerical solve.

\subsection{Stage Algorithm}
\label{sec:supp_stage_algorithm}

Algorithm~\ref{alg:supp_stage} summarizes one progressive stage. All regional
branches use the same segment boundaries and support budgets as the global
branch, so their target indices remain aligned for adaptive fusion.

\begin{algorithm}[t]
\caption{One AVIOT progressive compression stage}
\label{alg:supp_stage}
\begin{algorithmic}[1]
\REQUIRE Position-aware features $\mathbf X\in\mathbb R^{N\times S\times D}$,
question vector $\mathbf q_v$, target cardinality $K$
\STATE Form frame descriptors and split them into $M\leq4$ contiguous segments
\STATE Compute pilot couplings, segment relevance, and integer budgets
$\{k_m\}_{m=1}^{M}$
\FOR{each segment $m$}
    \STATE Initialize $k_m$ supports uniformly in time
    \STATE Alternate Eqs.~\eqref{eq:supp_damped_updates}--
    \eqref{eq:supp_barycentric_reconstruction} for the global descriptors
\ENDFOR
\STATE Concatenate segment plans in temporal order to obtain the global plan
\STATE Initialize and refine the nine medium plans from the global plan
\STATE Initialize and refine the 81 local plans from their medium parents
\STATE Apply each plan to its region and restore the $27\times27$ layout
\STATE Predict per-support, per-position fusion weights and combine the branches
\RETURN $\widetilde{\mathbf X}\in\mathbb R^{K\times S\times D}$ and the stage
feature-mixture coefficients
\end{algorithmic}
\end{algorithm}

\section{Architecture and Training Details}
\label{sec:supp_training}

This section specifies the AVIOT architecture, effective optimization
objective, and training schedule. Table~\ref{tab:supp_method_configuration}
summarizes its numerical configuration.

\begin{table*}[t]
    \centering
    \begingroup
    \small
    \setlength{\tabcolsep}{4pt}
    \begin{tabular*}{\textwidth}{@{\extracolsep{\fill}}lll@{}}
        \toprule
        Group & Setting & Value \\
        \midrule
        Representation & ViT feature dim. & 1152 \\
         & Patch layout & $27\times27$ \\
         & Cost embedding dim. & 256 \\
         & Temporal segments & 4 \\
        \midrule
        Transport & $\varepsilon$ & 0.10 \\
         & $\rho_s,\rho_t$ & 0.5, 5.0 \\
         & Log-domain updates & 20 \\
         & Global refinement rounds & 5 \\
         & Support initialization & Uniform temporal \\
        \midrule
        Progressive & Intermediate fractions & 0.75, 0.50, 0.25 \\
         & Cardinality rounding & Multiple of 8 \\
        \midrule
        Question & Allocation temperature & 1.0 $\rightarrow$ 0.1 (500 steps) \\
         & Segment-prior weight & 0.3 \\
        \midrule
        Spatial & Region blocks & 27, 9, 3 \\
         & Regions & 1, 9, 81 \\
         & Entropy coefficients & 0.10, 0.12, 0.15 \\
         & Executed refinement rounds (G/M/L) & 5 / 2 / 1 \\
         & Parent mixture (M/L) & 0.50 / 0.25 \\
        \midrule
        Fusion & Gate hidden dim. & 32 \\
         & Global floor & 0.20 \\
         & Temperature & 0.05 \\
        \midrule
        Position & THW Fourier bands & 16 per axis \\
         & Temporal/spatial scale & 1.0 / 0.5 \\
        \bottomrule
    \end{tabular*}
    \endgroup
    \caption{Final AVIOT architecture and numerical configuration.}
    \label{tab:supp_method_configuration}
\end{table*}

\subsection{Compression Objective}
\label{sec:supp_objective}

Suppose progressive stage $u$ contains temporal segments
$\{\mathcal I_m\}_{m=1}^{M_u}$. For a global segment, the visual-space
distortion is
\begin{equation}
\begin{aligned}
    \mathcal L_{\mathrm{temp},m}^{(u)}&=
    \frac{\sum_{i,j}\widehat P_{ij}^{(u),m}
    \|\mathbf x_i-\mathbf z_j\|_2^2/D}
    {\sum_{i,j}\widehat P_{ij}^{(u),m}},\\
    \mathcal L_{\mathrm{temp}}^{(u)}&=
    \sum_m\mathcal L_{\mathrm{temp},m}^{(u)}.
\end{aligned}
    \label{eq:supp_temporal_distortion}
\end{equation}
The coupling is determined in the 256-dimensional question-conditioned
transport space, whereas this distortion is measured between the corresponding
1152-dimensional source and barycentric target descriptors. Medium and local
regions use the same coupling-weighted visual-space distortion. Averaging their
respective losses gives
\begin{equation}
    \mathcal L_{\mathrm{reg}}^{(u)}=
    \tfrac12\left(\mathcal L_M^{(u)}+
    \mathcal L_L^{(u)}\right).
    \label{eq:supp_regional_distortion}
\end{equation}

Regional plans are allowed to differ, while visually similar adjacent regions
are encouraged to vary smoothly. Let $\mathcal E^\ell$ be the adjacent-region
pairs at granularity $\ell$. For $(r,r')\in\mathcal E^\ell$, let
$\mathbf A^{\ell,r}_{:,j}$ and $\mathbf A^{\ell,r'}_{:,j}$ be their
target-conditional distributions. Their weight is
\begin{equation}
    w_{rr'}^{\ell}=\exp\!\left[
    -\frac{1-\cos(\bar{\mathbf x}^{\ell,r},
    \bar{\mathbf x}^{\ell,r'})}{0.20}\right],
    \label{eq:supp_boundary_weight}
\end{equation}
where each $\bar{\mathbf x}$ is the normalized regional descriptor averaged
over source time. Here and below, $\operatorname{WAvg}$ normalizes the
displayed edge weights to sum to one and averages uniformly over the remaining
indices. The continuity term is
\begin{equation}
    \mathcal L_{\mathrm{cont}}^{(u)}=
    \frac12\sum_{\ell\in\{M,L\}}
    \operatorname{WAvg}^{w_{rr'}^\ell}_{(r,r')\in\mathcal E^\ell,j}
    \left[
    \operatorname{JS}(\mathbf A_{:,j}^{\ell,r},
    \mathbf A_{:,j}^{\ell,r'})\right].
    \label{eq:supp_continuity}
\end{equation}

The gate is regularized spatially with a motion-aware total-variation term.
Using the pre-compression visual features $\mathbf U$, the temporal change at
patch $s$ is
\begin{equation}
    d_s=\frac{1}{T-1}\sum_{i=2}^{T}
    [1-\cos(\mathbf U_{i,s},\mathbf U_{i-1,s})].
    \label{eq:supp_patch_motion}
\end{equation}
For adjacent patches $(s,s')$, define
$v_{ss'}=\exp(-|d_s-d_{s'}|/0.20)$, and let $\mathcal E$ collect all
horizontal and vertical patch pairs. The gate penalty is
\begin{equation}
    \mathcal L_{\mathrm{TV}}^{(u)}=
    \operatorname{WAvg}^{v_{ss'}}_{\ell,j,(s,s')\in\mathcal E}
    \left[|\gamma_{j,s}^{\ell}-\gamma_{j,s'}^{\ell}|\right].
    \label{eq:supp_gate_tv}
\end{equation}

Define the mean fusion weight of granularity $\ell$ over all target supports
and spatial positions as
$\bar\gamma^\ell=(K_uS)^{-1}\sum_{j,s}\gamma_{j,s}^{\ell}$. The anti-collapse
penalty is
\begin{equation}
\begin{aligned}
    \mathcal L_{\mathrm{bal}}^{(u)}=\frac13\big(&
    [\bar\gamma^G-0.60]_+^2+
    [0.15-\bar\gamma^M]_+^2\\
    &+[0.15-\bar\gamma^L]_+^2\big).
\end{aligned}
    \label{eq:supp_gate_balance}
\end{equation}
It imposes bounds rather than a fixed target mixture. A complementary entropy
barrier uses
\begin{equation}
\begin{aligned}
    H_\gamma^{(u)}&=-\frac{1}{K_uS\log3}
    \sum_{j,s,\ell}\gamma_{j,s}^{\ell}
    \log\gamma_{j,s}^{\ell},\\
    \mathcal L_{\mathrm{ent}}^{(u)}&=
    [0.85-H_\gamma^{(u)}]_+^2.
\end{aligned}
    \label{eq:supp_gate_entropy}
\end{equation}

The effective compression objective for one stage is
\begin{equation}
\begin{aligned}
    \mathcal L_{\mathrm{comp}}^{(u)}={}&
    \mathcal L_{\mathrm{temp}}^{(u)}
    +1.0\,\mathcal L_{\mathrm{reg}}^{(u)}
    +0.01\,\mathcal L_{\mathrm{cont}}^{(u)}\\
    &+0.001\,\mathcal L_{\mathrm{TV}}^{(u)}
    +0.5\,\mathcal L_{\mathrm{bal}}^{(u)}
    +0.5\,\mathcal L_{\mathrm{ent}}^{(u)}.
\end{aligned}
\label{eq:supp_stage_objective}
\end{equation}
For $J$ executed stages, this objective is averaged before it is added to the
answer-generation loss:
\begin{equation}
    \mathcal L=\mathcal L_{\mathrm{LM}}
    +\lambda_{\mathrm{OT}}\frac1J
    \sum_{u=1}^{J}\mathcal L_{\mathrm{comp}}^{(u)},
    \qquad \lambda_{\mathrm{OT}}=1.
    \label{eq:supp_total_objective}
\end{equation}
\subsection{Training Schedule and Model Components}
\label{sec:supp_training_schedule}

AVIOT is initialized from the LLaVA-Video-7B-Qwen2 checkpoint released with
LLaVA-Video~\citep{zhang2025llavavideo}. The SigLIP visual encoder is
frozen, while the language model, multimodal projector, THW and regional
position projections, question projection, transport metric, and fusion gate
are optimized jointly. The model is trained on LLaVA-Video-178K for 10,903
steps.

For every training example, one of the 17 ratios
$\{2,2.5,\ldots,9.5,10\}$ is sampled uniformly and converted to its target
cardinality with the ceiling rule in Sec.~\ref{sec:supp_progressive}. The
progressive stage sequence is then determined from that target. At inference,
the ratio is supplied externally and is not restricted to the training grid.
Table~\ref{tab:supp_training_configuration} gives the remaining optimization
settings.

\begin{table*}[t]
    \centering
    \begingroup
    \small
    \setlength{\tabcolsep}{4pt}
    \begin{tabular*}{\textwidth}{@{\extracolsep{\fill}}lll@{}}
        \toprule
        Group & Setting & Value \\
        \midrule
        Initialization & VideoLM & LLaVA-Video-7B-Qwen2 \\
         & Visual encoder & SigLIP-SO400M-patch14-384 \\
        \midrule
        Data & Training set & LLaVA-Video-178K \\
        \midrule
        Optimization & Steps & 10,903 \\
         & Global batch size & 128 \\
         & Learning rate & $1\times10^{-5}$ \\
         & Weight decay & 0 \\
         & Schedule & Cosine \\
         & Warmup ratio & 0.03 \\
        \midrule
        Compression & Training ratios & 2 to 10 in increments of 0.5 \\
         & Ratio sampling & Uniform over 17 choices \\
         & $\lambda_{\mathrm{OT}}$ & 1.0 \\
        \bottomrule
    \end{tabular*}
    \endgroup
    \caption{Optimization details for AVIOT training.}
    \label{tab:supp_training_configuration}
\end{table*}

\section{Extended Quantitative Results}
\label{sec:supp_results}

This section expands the aggregate results in the main paper with additional
benchmark breakdowns and compression-ratio analyses.
The evaluation covers Video-MME w/o sub., EgoSchema, MVBench, ActivityNet-QA,
NExT-QA, Perception Test, LongVideoBench, Video-MMMU, LVBench, and TempCompass.

\subsection{Controlled TempCompass Results}
\label{sec:supp_tempcompass_results}

\begin{table*}[t]
    \centering
    \begingroup
    \small
    \setlength{\tabcolsep}{4pt}
    \begin{tabular*}{\textwidth}{@{\extracolsep{\fill}}lcccccc@{}}
        \toprule
        Method & Ratio & Multi-choice & Yes/no & Caption matching & Captioning & Task avg. \\
        \midrule
        Uniform Keep & 2 & 67.15 & 68.94 & 75.45 & 53.29 & 66.21 \\
        Uniform Keep & 4 & 65.44 & 67.88 & 74.78 & 51.85 & 64.99 \\
        Uniform Keep & 10 & 63.54 & 65.80 & 72.46 & 12.33 & 53.53 \\
        \midrule
        Segment Mean & 2 & 66.52 & 68.98 & 75.32 & 53.54 & 66.09 \\
        Segment Mean & 4 & 64.81 & 66.82 & 73.72 & 51.50 & 64.21 \\
        Segment Mean & 10 & 60.32 & 63.72 & 68.80 & 14.07 & 51.73 \\
        \midrule
        AVIOT & 2 & \textbf{69.56} & \textbf{69.47} & \textbf{78.44} & \textbf{53.74} & \textbf{67.80} \\
        AVIOT & 4 & \textbf{68.16} & \textbf{68.69} & \textbf{78.24} & \textbf{53.09} & \textbf{67.05} \\
        AVIOT & 10 & \textbf{65.63} & \textbf{66.33} & \textbf{76.78} & \textbf{18.21} & \textbf{56.74} \\
        \bottomrule
    \end{tabular*}
    \endgroup
    \caption{Extended TempCompass results across four task types. Multi-choice
    follows the metric used in the main comparison; Task avg. is the unweighted
    mean of the four task accuracies. Bold marks the best compressed method at
    each ratio.}
    \label{tab:supp_tempcompass}
\end{table*}

Table~\ref{tab:supp_tempcompass} extends the controlled 64-frame comparison to
all four TempCompass tasks. The multi-choice column follows the metric used in
the main comparison, while the task average provides an additional unweighted
summary of the four task accuracies. AVIOT obtains the highest score for every
task and ratio among the three compressed methods. Its task average is 67.80,
67.05, and 56.74 at $r=2$, 4, and 10. Its margins over Uniform Keep are 1.59,
2.06, and 3.21 points, respectively, and its margins over Segment Mean are
1.71, 2.84, and 5.01 points. This breakdown resolves performance across
multiple-choice, yes/no, caption-matching, and captioning tasks.

The advantage is consistent across task types rather than being confined to a
single aggregate. Relative to the higher score produced by Uniform Keep or
Segment Mean for each task, AVIOT's margins at $r=2$ are 2.41 points on
multiple choice, 0.49 points on yes/no, 2.99 points on caption matching, and
0.20 points on captioning. At $r=4$, the corresponding margins are 2.72, 0.81,
3.46, and 1.24 points. At $r=10$, they are 2.09, 0.53, 4.32, and 4.14 points.
Thus the benefit over both simple inference-time baselines persists for both
recognition-style and temporally grounded generation-style questions as the
target cardinality is reduced.

The task average changes more gradually for AVIOT as the representation is
made progressively smaller. From $r=2$ to $r=10$, the average changes by
11.06 points for AVIOT, compared with 12.68 points for Uniform Keep and 14.36
points for Segment Mean. The gap consequently grows from 1.59 points at the
largest budget to 3.21 points at the smallest budget. This pattern indicates
that the advantage is not limited to the least compressed setting: the
transport-based construction continues to provide useful temporal evidence
when only a small number of supports is available.

\subsection{Performance Across Compression Ratios}
\label{sec:supp_ratio_results}

\begin{table*}[t]
    \centering
    \begingroup
    \scriptsize
    \setlength{\tabcolsep}{1.7pt}
    \begin{tabular*}{\textwidth}{@{\extracolsep{\fill}}lccccccccccccc@{}}
        \toprule
        Dataset & $r=2$ & $r=3.5$ & $r=4$ & $r=5$ & $r=5.5$ & $r=6$ & $r=6.5$ & $r=7$ & $r=8$ & $r=10$ & \shortstack{LLaVA-\\Video-7B} & CoPE \\
        \midrule
        Video-MME w/o sub. & 65.11 & 64.70 & 63.89 & 63.66 & 63.13 & 63.84 & 63.98 & 62.45 & 62.40 & 61.95 & 63.3 & 61.9 \\
        Video-MMMU & 38.22 & 39.00 & 39.78 & 38.33 & 38.00 & 38.78 & 38.78 & 37.78 & 38.33 & 37.89 & 36.1 & 38.2 \\
        ActivityNet-QA & 68.54 & 68.05 & 68.23 & 67.40 & 67.39 & 67.49 & 67.14 & 67.11 & 67.09 & 66.26 & 64.1 & 64.8 \\
        NExT-QA & 84.01 & 83.51 & 83.49 & 83.48 & 83.30 & 83.03 & 82.85 & 82.60 & 82.44 & 82.11 & 83.2 & 82.1 \\
        Perception Test & 71.84 & 71.48 & 71.45 & 71.03 & 70.86 & 70.57 & 70.52 & 70.48 & 70.27 & 69.80 & 67.9 & 70.3 \\
        TempCompass & 69.56 & 68.04 & 68.35 & 68.35 & 68.16 & 67.78 & 67.78 & 67.47 & 66.77 & 65.57 & 66.6 & 68.9 \\
        LongVideoBench & 58.19 & 57.37 & 56.99 & 56.62 & 57.14 & 56.10 & 55.72 & 55.50 & 55.72 & 53.78 & 58.2 & 56.9 \\
        LVBench & 45.53 & 44.74 & 45.38 & 44.74 & 43.13 & 43.59 & 42.80 & 42.41 & 42.50 & 40.87 & 44.2 & 46.4 \\
        EgoSchema & 58.04 & 57.34 & 58.14 & 57.58 & 57.50 & 57.32 & 57.26 & 57.64 & 57.05 & 56.47 & 57.3 & -- \\
        MVBench & 62.73 & 62.35 & 61.58 & 61.80 & 61.05 & 60.63 & 60.45 & 60.28 & 59.85 & 59.60 & 58.6 & 61.9 \\
        \bottomrule
    \end{tabular*}
    \endgroup
    \caption{AVIOT across compression ratios with 2 FPS sampling and at most
    224 frames. Scores are percentages, and TempCompass follows the
    multi-choice metric. The last two columns reproduce the published reference
    scores for LLaVA-Video-7B~\citep{zhang2025llavavideo} and
    CoPE~\citep{sarkar2026copevideolm}.}
    \label{tab:supp_ratio_results}
\end{table*}

Table~\ref{tab:supp_ratio_results} extends Table~2 of the main paper under the
same 2 FPS, at-most-224-frame setting. While Table~2 reports AVIOT at $r=4$,
Table~\ref{tab:supp_ratio_results} reports all evaluated ratios from $r=2$ to
$r=10$ under the same evaluation setting.

Across the ten benchmarks, the endpoint change from $r=2$ to $r=10$ is
dataset dependent, ranging from 0.33 points on Video-MMMU to 4.66 points on
LVBench. The corresponding decreases are 0.33 on Video-MMMU, 1.57 on
EgoSchema, 1.90 on NExT-QA, 2.04 on Perception Test, 2.28 on ActivityNet-QA,
3.13 on MVBench, 3.16 on Video-MME w/o sub., 3.99 on TempCompass, 4.41 on
LongVideoBench, and 4.66 on LVBench. The intermediate operating points provide
a numerical view of the accuracy--compression trade-off throughout the
training range. At $r=10$, AVIOT remains above the published LLaVA-Video-7B
reference~\citep{zhang2025llavavideo} on Video-MMMU (37.89 vs. 36.10),
ActivityNet-QA (66.26 vs. 64.10), Perception Test (69.80 vs. 67.90), and
MVBench (59.60 vs. 58.60). The additional high-ratio operating points are
listed in Table~\ref{tab:supp_high_ratio}.

\begin{table*}[t]
    \centering
    \begingroup
    \small
    \setlength{\tabcolsep}{6pt}
    \begin{tabular*}{\textwidth}{@{\extracolsep{\fill}}lcccccc@{}}
        \toprule
        Dataset & $r=12$ & $r=14$ & $r=15$ & $r=16$ & $r=18$ & $r=20$ \\
        \midrule
        Video-MME w/o sub. & 60.64 & 60.12 & 59.96 & 59.36 & 60.16 & 60.00 \\
        Video-MMMU & 36.44 & 36.89 & 36.22 & 36.89 & 36.56 & 36.78 \\
        MVBench & 58.13 & 58.10 & 58.00 & 57.90 & 57.75 & 57.38 \\
        NExT-QA & 81.71 & 81.90 & 81.80 & 81.66 & 81.48 & 81.26 \\
        Perception Test & 69.17 & 69.00 & 68.86 & 68.77 & 68.48 & 68.30 \\
        \bottomrule
    \end{tabular*}
    \endgroup
    \caption{AVIOT at inference-time compression ratios beyond the training
    range for the five benchmarks in the main-paper ratio analysis, using 2 FPS
    sampling and at most 224 frames. All operating points use the same
    checkpoint and target-cardinality construction as the main results.}
    \label{tab:supp_high_ratio}
\end{table*}

Table~\ref{tab:supp_high_ratio} reports the numerical results at inference-time
compression ratios beyond the training range for the five benchmarks shown in
the main paper's compression-ratio analysis. Between $r=12$ and $r=20$, the
endpoint changes are 0.64 points on Video-MME w/o sub., 0.34 on Video-MMMU,
0.75 on MVBench, 0.45 on NExT-QA, and 0.87 on Perception Test. Relative to
$r=10$, the decreases at $r=20$ are 1.95, 1.11, 2.22, 0.85, and 1.50 points,
respectively. These operating points show that the same checkpoint continues
to provide direct inference-time control of target cardinality at up to twice
the largest training ratio.

Together, the two tables provide a common ratio sweep for all ten benchmarks
throughout the training range and exact out-of-training-range scores for five
of them. At $r=10$, the ten
benchmark scores retain 89.8--99.1\% of their respective $r=2$ values, and
eight of the ten retain more than 93\%. From $r=10$ to $r=20$, the additional
decrease on the five reported benchmarks remains at most 2.22 points. These
results characterize a broad set of operating points at which the same
checkpoint trades target cardinality for compression without being tied to a
single inference-time ratio.

\subsection{Robustness Across Input Sampling}
\label{sec:supp_sampling_robustness}

\begin{table*}[t]
    \centering
    \begingroup
    \scriptsize
    \setlength{\tabcolsep}{4pt}
    \begin{tabular*}{\textwidth}{@{\extracolsep{\fill}}lcccccc@{}}
        \toprule
        Dataset & \shortstack{64 frames\\$r=2$} &
        \shortstack{2 FPS, $\leq$224\\$r=2$} &
        \shortstack{64 frames\\$r=4$} &
        \shortstack{2 FPS, $\leq$224\\$r=4$} &
        \shortstack{64 frames\\$r=10$} &
        \shortstack{2 FPS, $\leq$224\\$r=10$} \\
        \midrule
        ActivityNet-QA & 68.22 & 68.54 & 67.06 & 68.23 & 65.67 & 66.26 \\
        Perception Test & 71.50 & 71.84 & 70.72 & 71.45 & 69.41 & 69.80 \\
        MVBench & 62.20 & 62.73 & 61.33 & 61.58 & 58.65 & 59.60 \\
        Video-MME w/o sub. & 63.22 & 65.11 & 60.85 & 63.89 & 58.26 & 61.95 \\
        EgoSchema & 57.86 & 58.04 & 56.33 & 58.14 & 53.61 & 56.47 \\
        \bottomrule
    \end{tabular*}
    \endgroup
    \caption{AVIOT under fixed-64-frame and 2 FPS, at most 224-frame input
    sampling. Scores are reported at compression ratios 2, 4, and 10.}
    \label{tab:supp_videomme_duration}
\end{table*}

Table~\ref{tab:supp_videomme_duration} compares AVIOT at the same ratios when
the input is either fixed to 64 frames or sampled at 2 FPS with at most 224
frames. The results show that the compact representation remains effective
under both input-sampling choices, while denser temporal input can provide
additional evidence when it is available.

Relative to fixed 64-frame input, using the denser sampling
raises ActivityNet-QA by 0.32, 1.17, and 0.59 points at ratios 2, 4, and 10;
Perception Test by 0.34, 0.73, and 0.39 points; and MVBench by 0.53, 0.25, and
0.95 points. Video-MME w/o sub. gains 1.89, 3.04, and 3.69 points at the same ratios,
while EgoSchema gains 0.18, 1.81, and 2.86 points at ratios 2, 4, and 10. For ActivityNet-QA,
Perception Test, MVBench, and Video-MME w/o sub., every displayed pair favors the denser
input, with the largest gain occurring on Video-MME w/o sub. at $r=10$. The resulting
differences are modest relative to the compression effect itself: averaged over
the five datasets, the gain is 0.65, 1.40, and 1.70 points at $r=2$, 4, and
10. AVIOT therefore benefits from additional source observations when they are
available without relying on one particular input-frame count.

\subsection{MVBench Task Breakdown}
\label{sec:supp_mvbench_breakdown}

\begin{table*}[t]
    \centering
    \begingroup
    \small
    \setlength{\tabcolsep}{4.2pt}
    \begin{tabular*}{\textwidth}{@{\extracolsep{\fill}}lccclccc@{}}
        \toprule
        Task & $r=2$ & $r=4$ & $r=10$ & Task & $r=2$ & $r=4$ & $r=10$ \\
        \midrule
        Action antonym & 79.50 & 78.00 & 72.50 & Fine-grained pose & 58.50 & 57.00 & 49.50 \\
        Action count & 59.50 & 60.00 & 58.50 & Moving attribute & 68.00 & 68.50 & 64.50 \\
        Action localization & 63.50 & 60.50 & 57.50 & Moving count & 46.00 & 45.50 & 44.50 \\
        Action prediction & 66.00 & 64.00 & 60.50 & Moving direction & 37.00 & 33.50 & 29.50 \\
        Action sequence & 77.00 & 75.50 & 70.50 & Object existence & 58.00 & 58.50 & 57.50 \\
        Character order & 79.00 & 80.00 & 72.50 & Object interaction & 82.00 & 79.50 & 80.00 \\
        Counterfactual inference & 51.50 & 51.00 & 48.50 & Object shuffle & 40.50 & 41.00 & 38.50 \\
        Egocentric navigation & 38.00 & 37.00 & 36.50 & Scene transition & 93.00 & 91.50 & 93.00 \\
        Episodic reasoning & 54.50 & 54.50 & 53.00 & State change & 63.50 & 64.50 & 62.50 \\
        Fine-grained action & 47.50 & 46.00 & 45.50 & Unexpected action & 81.50 & 80.50 & 78.00 \\
        \midrule
        Overall & 62.20 & 61.33 & 58.65 & \multicolumn{4}{c}{} \\
        \bottomrule
    \end{tabular*}
    \endgroup
    \caption{AVIOT accuracy on the 20 MVBench tasks at three compression ratios.}
    \label{tab:supp_mvbench_tasks}
\end{table*}

Table~\ref{tab:supp_mvbench_tasks} further resolves the controlled AVIOT result
over all 20 MVBench tasks. Increasing the ratio from 2 to 10 leaves 13 tasks
within five points and 19 tasks within eight points of their $r=2$ accuracy.
Scene transition is unchanged at 93.0, while object existence, action count,
state change, and several reasoning tasks change by at most three points. This
task-level view shows that AVIOT retains strong performance across diverse task
categories as the target cardinality is reduced.

The task-level changes also reveal which abilities make the greatest use of
additional temporal resolution. Object existence, object interaction, state change,
counterfactual inference, episodic reasoning, and the two counting tasks remain
within three points of their $r=2$ scores. The larger changes occur on action
antonym, action localization, action sequence, character order, fine-grained
pose, and moving direction, which require finer temporal ordering or motion
discrimination. The resulting pattern is consistent with the role of the
compressed supports: broad event and state evidence remains stable, while
fine temporal distinctions benefit most from retaining additional supports.

The intermediate setting reinforces this separation. The overall score changes
from 62.20 at $r=2$ to 61.33 at $r=4$ and 58.65 at $r=10$, while several
state- and object-oriented tasks are nearly unchanged between the first two
settings. In contrast, the strongest ratio-dependent variations at $r=10$ are
concentrated in motion direction, fine-grained pose, and action ordering. The task breakdown
therefore connects the aggregate ratio curve to the type of evidence retained
by each support: the construction preserves broad scene and state cues while
providing a controllable temporal-resolution trade-off for fine motion
distinctions.

\section{Visual-Token and Decoder-Prefill Accounting}
\label{sec:supp_efficiency}

AVIOT reduces the temporal cardinality of the visual representation before it
enters language-model decoding. This section makes the visual-token and
decoder-prefill accounting in Table~1 of the main paper explicit. The quantities
below isolate the part of inference directly changed by compression: the visual
prefix consumed by the language model after the source representation has been
constructed.

\subsection{Visual-Token Accounting}

Let $T$ be the number of encoded source frames and
$K=\lceil T/r\rceil$ the requested number of target supports. AVIOT preserves
the $27\times27$ spatial layout while constructing these supports. The inherited
multimodal path projects each support to the language-model dimension and
bilinearly pools its layout to $14\times14$. Grid formatting then appends one
newline embedding to each of the 14 rows. Consequently, the number of visual
tokens presented to the language model is exactly
\begin{equation}
    n_{\mathrm{vis}}(K)
    =K(14^2+14)=210K.
    \label{eq:supp_visual_tokens}
\end{equation}
The token-retention fraction relative to the uncompressed representation is
therefore $K/T$. For $T=64$, ratios 2, 4, and 10 give
$K=32,16,$ and $7$, respectively.

\subsection{Decoder-Prefill Accounting}

The prefill values use the same architecture-level convention for every method
in the controlled table. For a decoder with $L$ layers, hidden dimension $d$,
feed-forward dimension $d_{\mathrm{ff}}$, and a visual prefix of $n$ tokens, the
count is
\begin{equation}
    F_{\mathrm{prefill}}(n)
    =L\left(4nd^2+2n^2d+2ndd_{\mathrm{ff}}\right).
    \label{eq:supp_prefill_flops}
\end{equation}
The three terms account for attention projections, causal self-attention, and
feed-forward projections under this common analytical convention. It excludes
embedding lookup, normalization, elementwise operations, the output head, and
prompt-dependent text tokens. The final backbone has $L=28$, $d=3584$, and
$d_{\mathrm{ff}}=18944$. Substituting Eq.~\eqref{eq:supp_visual_tokens} gives
the values in Table~\ref{tab:supp_complexity}.

\begin{table}[t]
    \centering
    \begingroup
    \small
    \setlength{\tabcolsep}{4.0pt}
    \begin{tabular}{lrrrr}
        \toprule
        Setting & $K$ & Visual tokens & Retention & Prefill (T) \\
        \midrule
        Uncompressed & 64 & 13,440 & 100.0\% & 106.69 \\
        $r=2$ & 32 & 6,720 & 50.0\% & 44.28 \\
        $r=4$ & 16 & 3,360 & 25.0\% & 19.87 \\
        $r=10$ & 7 & 1,470 & 10.9\% & 8.14 \\
        \bottomrule
    \end{tabular}
    \endgroup
    \caption{Visual-token and decoder-prefill accounting for the controlled 64-frame setting. Prefill follows Eq.~\eqref{eq:supp_prefill_flops}.}
    \label{tab:supp_complexity}
\end{table}

At $r=2$, 4, and 10, the standardized prefill count is reduced by 58.5\%,
81.4\%, and 92.4\% relative to the 64-frame visual prefix. These values are
defined at the decoder interface, where compression changes the length of the
visual prefix. At a fixed target cardinality, Uniform Keep, Segment Mean, and
AVIOT present the same $210K$-token layout to the inherited multimodal path;
their distinction is how those $K$ supports are constructed.

\section{Additional Qualitative Analysis}
\label{sec:supp_qualitative}

This section provides controlled visual evidence for question-conditioned
temporal allocation and transport across spatial granularities, followed by
three additional provenance cases spanning distinct video types. The displayed
quantities are exact feature-mixture coefficients obtained by composing the
model's target-conditional distributions and adaptive fusion weights across
progressive compression stages. For each target support and spatial position,
the coefficients sum over the original source frames and reconstruct the
corresponding compressed feature up to numerical precision; they are not
post-hoc attention or saliency estimates.

\subsection{Question-Conditioned Temporal Allocation}

\begin{figure*}[!b]
    \centering
    \includegraphics[width=\textwidth]{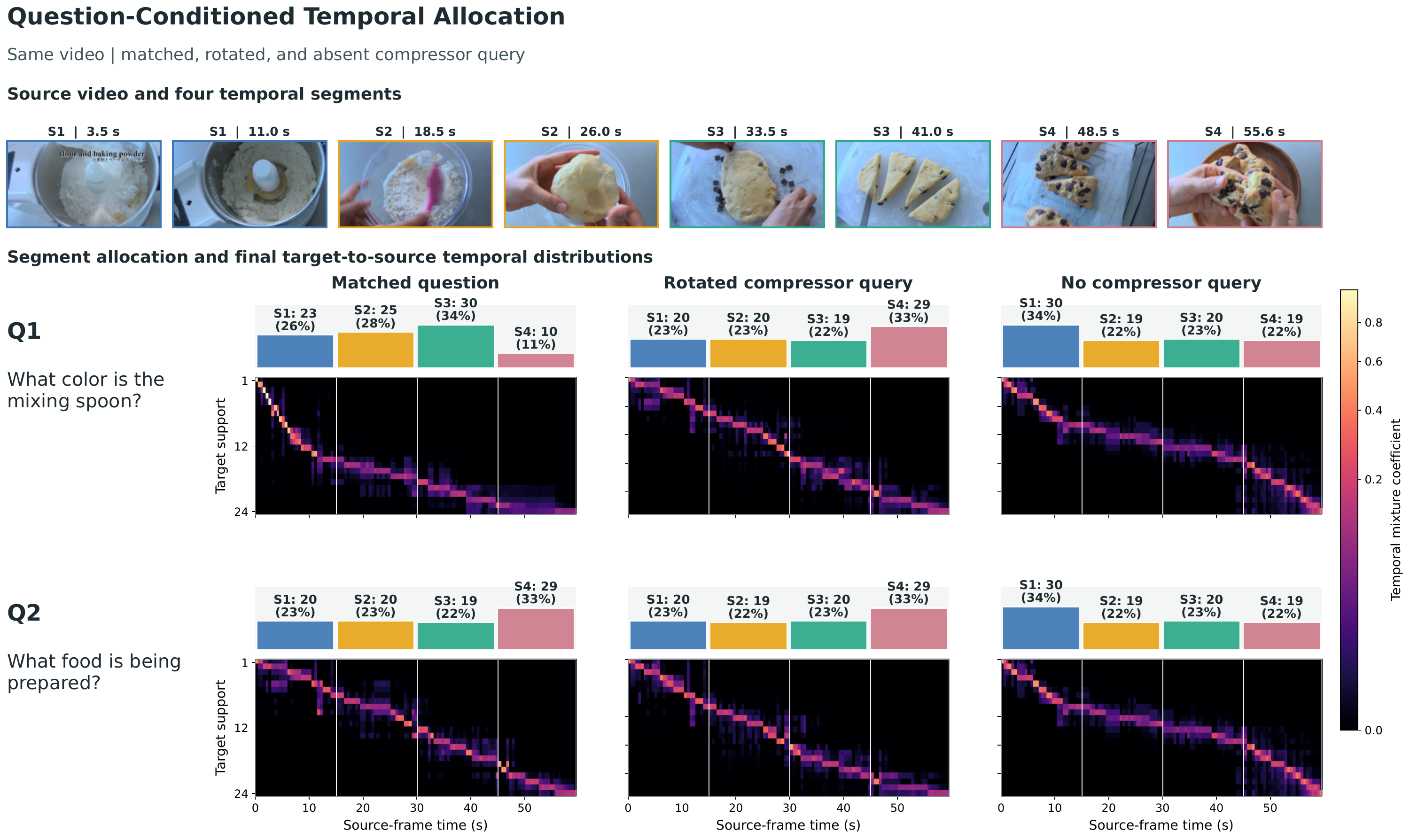}
    \caption{Controlled question-conditioned temporal allocation on a fixed
    video and support budget. Bar heights encode the support share assigned to
    each temporal segment at the first progressive stage; heatmap rows are the
    exact source-time distributions of ordered final target supports. The
    matched column supplies the displayed question to AVIOT, the rotated-query
    column supplies a different fixed question from the same video, and the
    no-query column disables all compressor-side question routes.}
    \label{fig:supp_question_allocation}
\end{figure*}

Figure~\ref{fig:supp_question_allocation} compares two questions about the same
119-frame video while fixing the compression ratio, temporal segmentation, and
final cardinality of 24 target supports ($r=5$). The bars report the support
allocation at the first progressive stage, and the heatmaps show the exact
source-time distributions of the ordered final supports. We evaluate three
compressor-side conditions while always retaining the correct displayed
question in the language-model prompt. In the \emph{matched} condition, the
compressor receives that same question. In the \emph{rotated-query} condition,
it instead receives a different, fixed question from the same video. In the
\emph{no-query} condition, question input is disabled for all compressor routes,
including support allocation, temporal and regional transport, and adaptive
fusion.

Under matched conditioning, the spoon-color question allocates
$[23,25,30,10]$ supports to the four segments, placing 88.6\% of the first
stage's budget in the first three segments that contain the spoon and mixing
action. The food-type question instead allocates $[20,20,19,29]$, increasing
the final, food-revealing segment from 11.4\% to 33.0\% of the budget. To
quantify this change, we normalize each four-segment allocation to a vector
$\mathbf a$ that sums to one and compute
$\lVert\mathbf a^{\mathrm{Q1}}-\mathbf a^{\mathrm{Q2}}\rVert_1$. We also
average the 24 row-normalized heatmap distributions over target supports and
compute the Jensen--Shannon divergence between the resulting source-time
distributions. The allocation distance ranges from 0 to 2, and the temporal
divergence from 0 to 1 bit. Each is zero when the corresponding allocations or
temporal distributions are identical; larger values indicate greater
between-question separation. The matched condition gives an allocation
distance of 0.432, equivalent to reallocating 21.6\% of the stage budget
between segments, and a temporal divergence of 0.155 bits.

The two controls show that this separation is caused by the question supplied
to the compressor. With rotated compressor queries, the allocations become
$[20,20,19,29]$ and $[20,19,20,29]$: they differ by only one support exchanged
between the second and third segments, reducing the allocation distance to
0.023 and the temporal divergence to 0.022 bits. With no compressor query, both
questions produce the same allocation, $[30,19,20,19]$, and identical complete
feature-mixture coefficient tensors, including the displayed target-to-source
matrices, so both measures are zero. Because the video, compression setting,
and correct language-model prompts remain fixed, these interventions directly
verify that question conditioning changes both the temporal allocation of
representation capacity and the source-time composition of the final supports.

\subsection{Transport Across Spatial Granularities}

\begin{figure*}[!t]
    \centering
    \includegraphics[width=\textwidth]{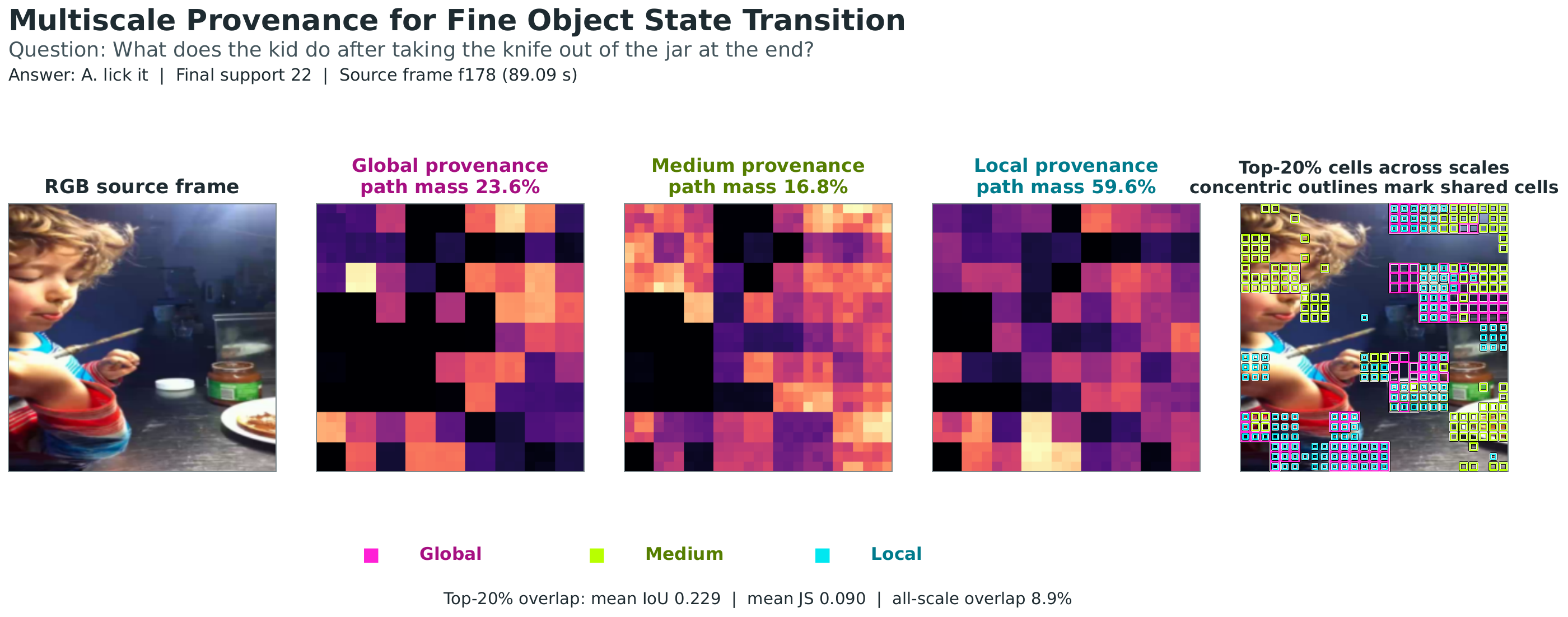}
    \caption{Exact provenance across spatial granularities for one source frame
    and final target support. Each branch map is normalized independently to
    expose its spatial pattern; the percentages report its contribution mass
    after adaptive fusion. Colored outlines in the rightmost panel mark the
    top-20\% cells selected by the global, medium, and local paths, with
    concentric outlines indicating cross-path overlap.}
    \label{fig:supp_spatial_granularities}
\end{figure*}

Figure~\ref{fig:supp_spatial_granularities} examines a fine object-state
transition for the question ``What does the kid do after taking the knife out
of the jar at the end?'' The correct answer is ``lick it.'' For final target
support 22 and source frame 178 ($89.09\,\mathrm{s}$), the global, medium, and
local paths contribute 23.6\%, 16.8\%, and 59.6\% of the fused provenance mass,
respectively. Although the local path contributes the largest share, the
global and medium paths jointly retain 40.4\% of the contribution for this
source--target pair.

The three paths first exhibit a common spatial focus. After normalizing each
branch map to unit mass and expressing distance relative to the image
diagonal, their mass-weighted centroids have a mean pairwise distance of 0.039
and a maximum distance of 0.056. They therefore remain centered on the same
child and hand--object interaction rather than drifting to unrelated regions.
Within this shared focus, however, the independently normalized maps do not
repeat one spatial pattern. Their top-20\% cell sets have a mean pairwise IoU
of 0.229 and a mean Jensen--Shannon divergence of 0.090, while only 8.9\% of
one branch's top-20\% set is present in the three-way intersection. The
highlighted regions cover different combinations of the child's broader
context, the hand--object relation, and local evidence around the knife and
face. The aligned centroids and distinct internal distributions together show
how the three granularities retain a common event through complementary
spatial contexts rather than replicating one correspondence pattern.

\subsection{Exact Provenance Across Video Types}

Three further cases cover a long first-person video, a state transition, and a
fine-grained tool interaction. Each curve plots $\omega_i$, the spatially
averaged provenance coefficient for one target support, over the displayed
source-frame interval; the purple marker identifies the displayed frame. Yellow
grids mark the top-20\% cells, and the $27\times27$ map shows the complete
coefficient field. As supports are constructed within temporal segments, each
curve and map shows contributions within its segment.

\begin{figure*}[p]
    \centering
    \includegraphics[width=\textwidth]{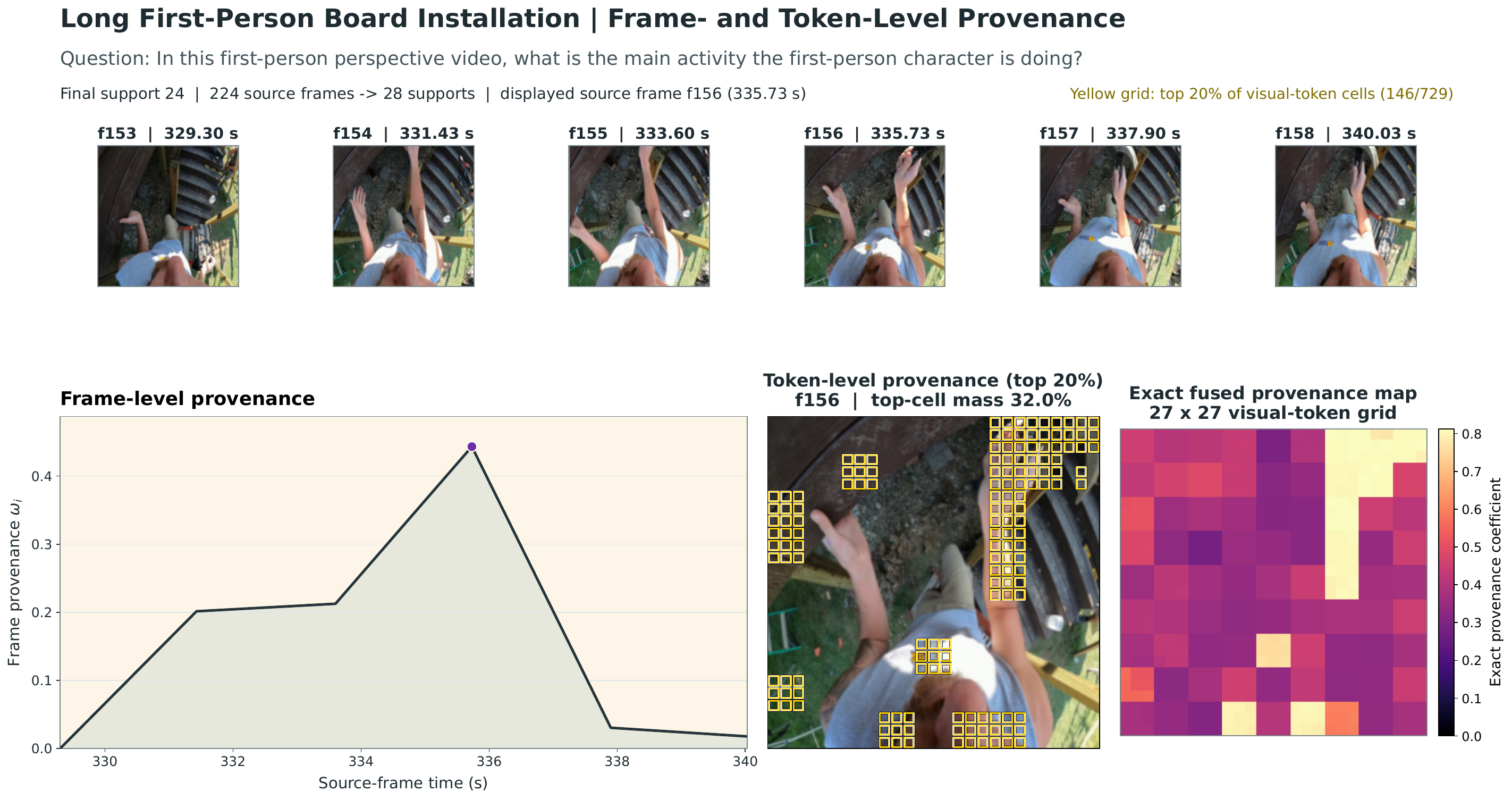}
    \caption{Exact frame- and token-level provenance for a target support
    in a long first-person board-installation video. The temporal curve reports
    $\omega_i$ over the displayed source-frame interval. The token-level views
    show the complete spatial mixture for the displayed source frame.}
    \label{fig:supp_board_provenance}
\end{figure*}

Figure~\ref{fig:supp_board_provenance} traces target support 24 in an
eight-minute, 224-frame video compressed to 28 supports. Frames 153--158 follow
the hands and board through successive installation views. At displayed frame 156
($335.73\,\mathrm{s}$), the top-20\% cells carry 32.0\% of the fused
provenance mass over the hands, board, and work area; the complete map retains
broader context, combining temporal development with actor--object evidence.

\begin{figure*}[p]
    \centering
    \includegraphics[width=\textwidth]{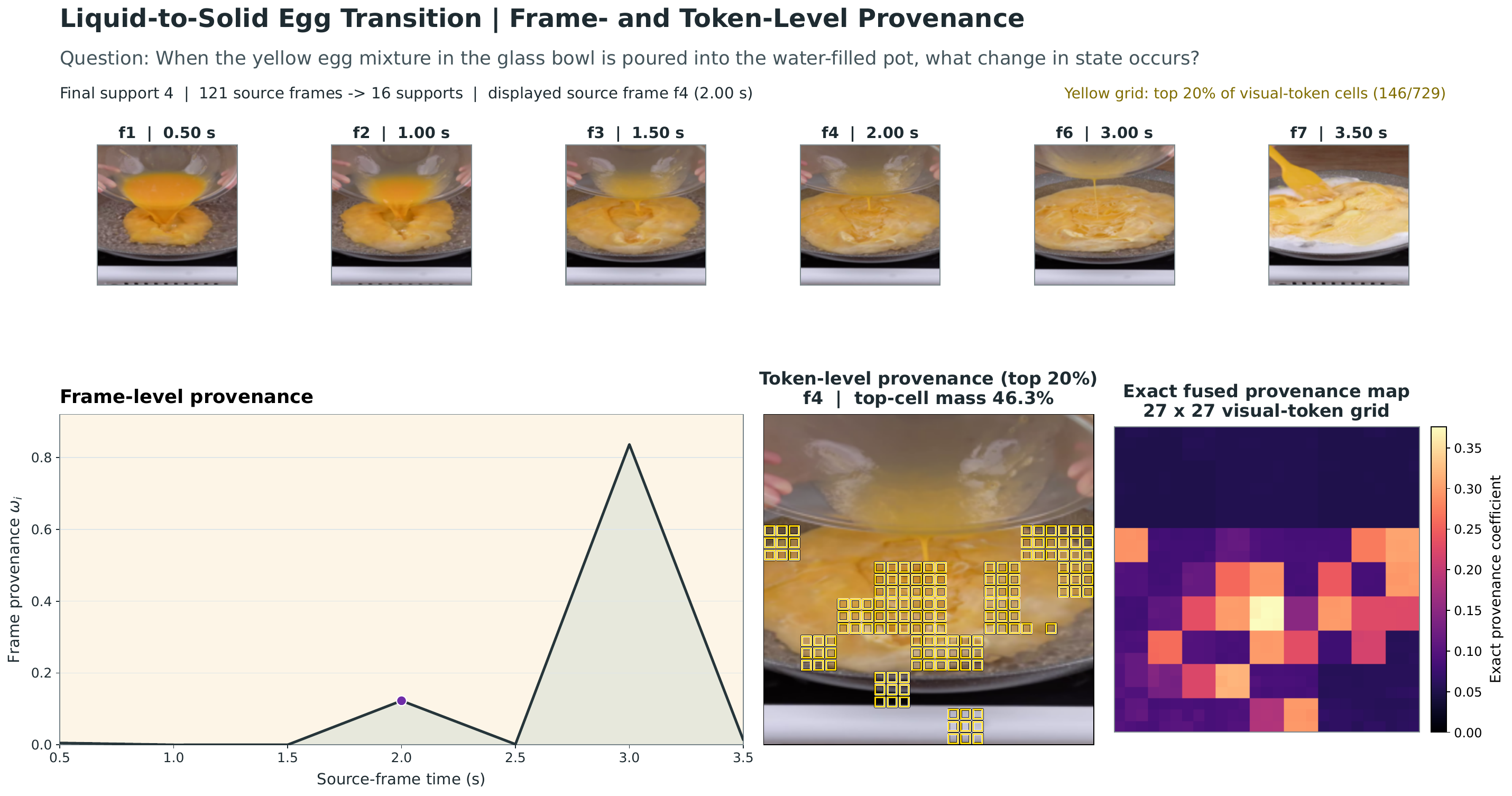}
    \caption{Exact provenance for a target support spanning the liquid-to-solid
    egg transition. The frame-level curve reports $\omega_i$ over the displayed
    source-frame interval. The token-level views show its spatial construction
    within the displayed source frame.}
    \label{fig:supp_egg_provenance}
\end{figure*}

Figure~\ref{fig:supp_egg_provenance} examines a
liquid-to-solid transition in a 121-frame video compressed to 16 supports.
For support 4, frames 1--7 span pouring to the set state. At
displayed frame 4 ($2.00\,\mathrm{s}$), the top-20\% cells carry 46.3\% of the
spatial provenance mass over the mixture, pot interior, and boundary; the
complete map retains surrounding context and locates the transition spatially.

\begin{figure*}[p]
    \centering
    \includegraphics[width=\textwidth]{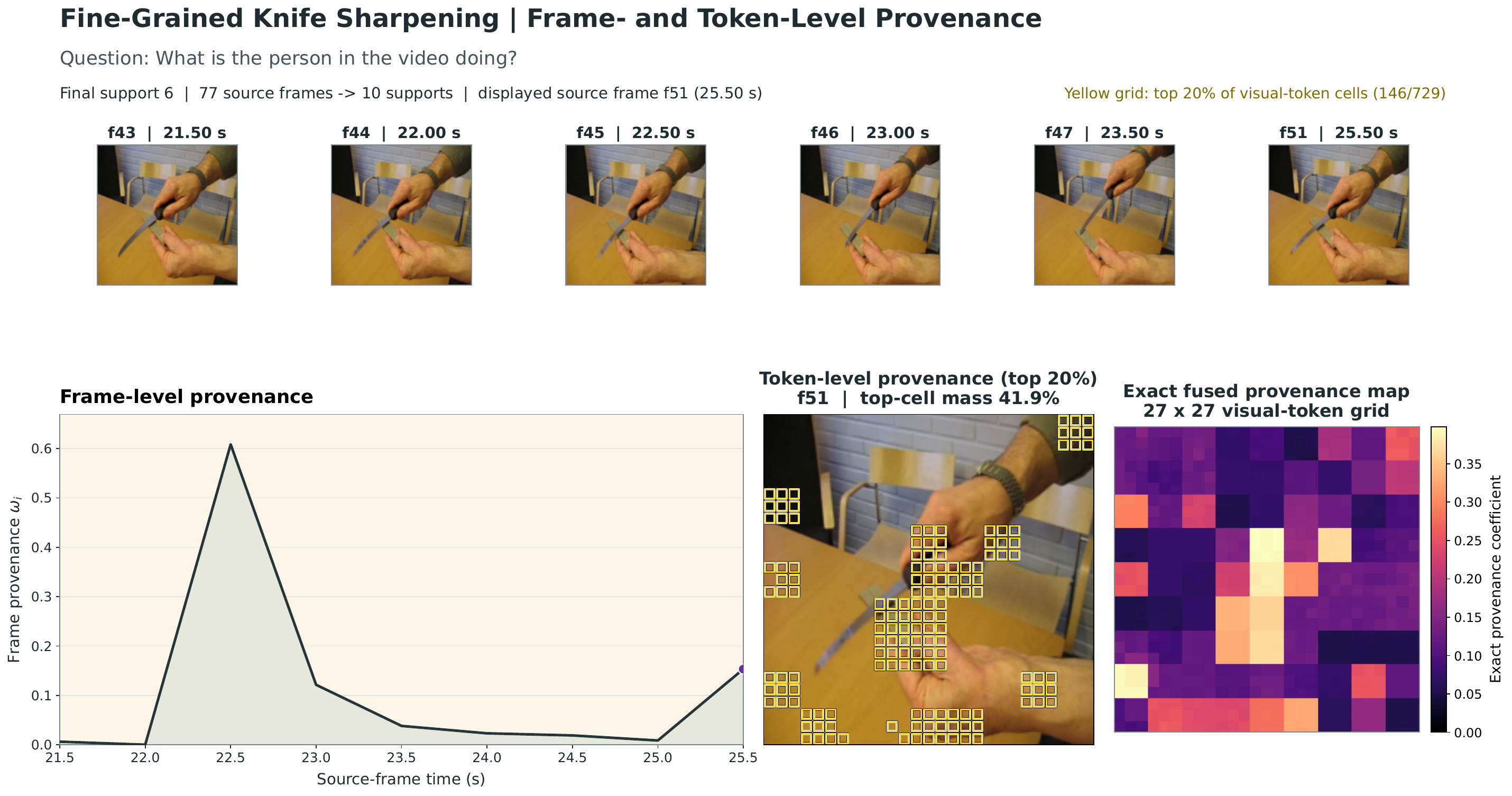}
    \caption{Exact provenance for a target support representing fine-grained
    knife sharpening. The temporal curve reports $\omega_i$ over the displayed
    source-frame interval. The token-level views preserve the geometry of the
    hands, knife, and sharpening tool within the displayed source frame.}
    \label{fig:supp_knife_provenance}
\end{figure*}

Figure~\ref{fig:supp_knife_provenance} captures knife sharpening in a 77-frame
video compressed to 10 supports. For support 6, frames 43--47 and 51 show
successive and repeated sharpening states. At displayed frame 51
($25.50\,\mathrm{s}$), the
top-20\% cells carry 41.9\% of the spatial provenance mass over the hands,
knife, sharpening tool, and contact region; the complete field retains the
scene context that distinguishes sharpening from generic tool use.

Across the three cases, segment-wise provenance describes localized,
transitional, and repeated interactions while retaining broader context.
Together with the controlled query intervention and branch-specific
maps, these cases show how AVIOT adapts representation capacity, resolves
correspondence across spatial contexts, and integrates distributed observations
into compact representations.

\section{Limitations}
\label{sec:supp_limitations}

AVIOT is applied after the sampled frames have been encoded. It reduces the
number of visual tokens passed to the language model, thereby lowering
language-model prefill cost, but does not reduce vision-encoder computation. Our
experiments use LLaVA-Video-7B-Qwen2 and a regular $27\times27$ SigLIP feature
grid; evaluating the method with other language-model backbones, vision
encoders, spatial layouts, and streaming video is left to future work. The
compression ratio can be specified as needed at inference time, whereas
automatically selecting a video- and question-dependent ratio remains future
work.

\bibliographystyle{aviotpaper}
\bibliography{references}